\documentclass[10pt,twocolumn,letterpaper]{article}

\usepackage{cvpr}              
\usepackage[none]{hyphenat}

\usepackage{graphicx}
\usepackage{multirow}
\usepackage{booktabs}
\usepackage[table,xcdraw]{xcolor}
\usepackage{makecell}
\definecolor{cvprblue}{rgb}{0.21,0.49,0.74}
\usepackage[pagebackref,breaklinks,colorlinks,allcolors=cvprblue]{hyperref}

\def\paperID{***} 
\def\confName{CVPR}
\def\confYear{2026}

\title{
Learning to See Through a Baby’s Eyes:\\ Early Visual Diets Enable Robust Visual Intelligence in Humans and Machines
}

\author{Yusen Cai$^{1}$
\quad
Qing Lin$^{1,*}$
\quad
Bhargava Satya Nunna$^{1,2}$
\quad
Mengmi Zhang$^{1,*}$ 
\\
$^{1}$Nanyang Technological University, Singapore \quad
$^{2}$Indian Institute Of Technology Madras\\
$^{*}$Co-corresponding authors\\
{\tt \small \{qing.lin, mengmi.zhang\}@ntu.edu.sg} 
}

\begin{document}

\maketitle
\begin{abstract}
Newborns perceive the world with low-acuity, color-degraded, and temporally continuous vision, which gradually sharpens as infants develop. To explore the ecological advantages of such staged ``visual diets", we train self-supervised learning (SSL) models on object-centric videos under constraints that simulate infant vision: grayscale-to-color (C), blur-to-sharp (A), and preserved temporal continuity (T)—collectively termed CATDiet. For evaluation, we establish a comprehensive benchmark across ten datasets, covering clean and corrupted image recognition, texture–shape cue conflict tests, silhouette recognition, depth-order classification, and the visual cliff paradigm.
All CATDiet variants demonstrate enhanced robustness in object recognition, despite being trained solely on object-centric videos. Remarkably, models also exhibit biologically aligned developmental patterns, including neural plasticity changes mirroring synaptic density in macaque V1 and behaviors resembling infants’ visual cliff responses. Building on these insights, CombDiet initializes SSL with CATDiet before standard training while preserving temporal continuity. Trained on object-centric or head-mounted infant videos, CombDiet outperforms standard SSL on both in-domain and out-of-domain object recognition and depth perception. Together, these results suggest that the developmental progression of early infant visual experience offers a powerful reverse-engineering framework for understanding the emergence of robust visual intelligence in machines. All code, data, and models are available at \href{https://github.com/ZhangLab-DeepNeuroCogLab/CATDiet}{Github}. 
\end{abstract}
\section{Introduction}

\begin{figure*}[t] 
    \centering
    \includegraphics[width=\textwidth]{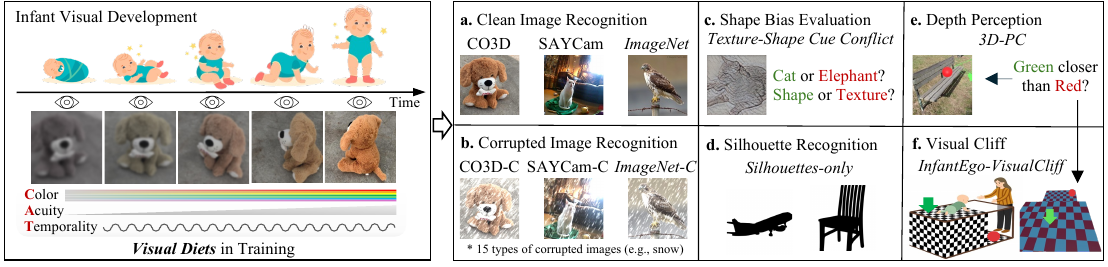}
    \vspace{-4mm}
    \caption{\textbf{Illustration of developmental visual diets and overview of evaluation benchmarks.} 
The \textbf{left} panel depicts stages of infant development over time along with corresponding characteristics of visual perception. The 3 axes below represent the key regularities of infant visual development that underpin our work: Color, Acuity, and Temporality. The \textit{Color} diet (CDiet) models the progression from color-degraded to richly chromatic scenes as color vision matures. The \textit{Acuity} diet (ADiet) reflects the transition from blurry to sharp perception as visual resolution improves. 
The \textit{Temporality} diet (TDiet) captures infants’ exposure to smoothly evolving visual scenes over short time windows.
    In the \textbf{right} panel, regular font indicates in-domain tasks, while \textit{italic font} denotes out-of-domain tasks. 
Panels (a–d) assess object recognition: 
(a) clean image recognition on CO3D \cite{reizenstein21co3d}, SAYCam \cite{orhan2020self}, and ImageNet \cite{ridnik2021in21k}; 
(b) corrupted image recognition across 15 corruption types following ImageNet-C \cite{hendrycks2019benchmarking}; 
(c) shape-bias evaluation using the Texture–Shape Cue Conflict dataset \cite{geirhos2018imagenet}, testing whether classification aligns with shapes (green) or textures (red); and 
(d) silhouette recognition using the Silhouettes-Only dataset \cite{geirhos2018imagenet}. 
Panels (e–f) evaluate depth perception: 
(e) judging whether the green arrow is closer than the red ball in the 3D-PC dataset \cite{linsley20243d}; and 
(f) predicting which side (green arrow or red ball) appears closer from the infant egocentric perspective in the Visual Cliff paradigm \cite{gibson1960viscliff}.
    } 
\vspace{-4mm}
    \label{fig:illustration}
\end{figure*}

Newborns perceive a world that is blurry, desaturated, and continuously unfolding in time \cite{vogelsang2024impact, maule2023development, werner1979human, vogelsang2018potential, vogelsang2025potential, wood2018development}. Over the first year of life, visual acuity sharpens and color sensitivity emerges as their visual inputs mature \cite{norcia1985spatial, aslin2001visual, mayer1982visual, teller1997first, skelton2022infant, crognale2002development}. This early limitation is not a defect but a developmental scaffold, enabling the brain to organize sensory experience into stable and generalizable representations \cite{zaadnoordijk2022lessonsnature, vogelsang2024butterfly, smith2005development, smith2018developing, smith2017developmental, bambach2018toddler, blakemore1970development, dekker2025infants, dominguez2003binocular}. When this developmental process is disrupted, perceptual deficits can arise. For instance, the absence of early low-acuity input has been linked to lasting deficits in face processing \cite{vogelsang2018potential, geldart2002effect, maurer2007sleeper, putzar2010early}, and immature photoreceptors bias infants toward luminance-based rather than chromatic cues \cite{vogelsang2024impact}.

In contrast, artificial visual systems in AI are typically trained on fully detailed, static images. To increase data diversity and improve model robustness, random data augmentations are commonly applied \cite{yun2019cutmixrf12,hendrycks2019augmixrf13,zhang2017mixuprf14,he2022masked, rebuffi2021data, zhong2020random, chen2020gridmask, cubuk2019autoaugment}. While such training regimes achieve impressive performance on standard benchmarks, they remain ecologically invalid, overlooking how perception in nature develops through structured and temporally coherent visual experience \cite{talbot2023tuned, tee2023integrating, singh2023learning, wu2023label, madan2022improving}. As a result, modern vision models struggle to generalize to corrupted or occluded images \cite{bomatter2021pigs,zhang2020putting,liu2022reason,dodge2016quality, hendrycks2019benchmarking, vasiljevic2016blur, wang2023survey, mintun2021interaction, singh2024synthetic, wang2024unsupervised, chiu2022colormachine}. This limitation constrains their reliability in high-stakes real-world applications, including autonomous driving, robotics, and embodied human–AI interaction \cite{khandelwal2023adaptive,zhang2025peering, zhang2024robot2, nguyen2021robot1, xie2025driving2, wang2024driving1, zhao2024driving3, xu2024embodied1, yang2025embodied2}.

What enables human perceptual robustness, and how might we reverse-engineer it? Decades of developmental research suggest two key factors. First, the structure of visual experience — including spatial statistics, infant-perspective object distributions, and temporal continuity — guides learning toward generalizable shape-based representations rather than brittle cue-specific solutions \cite{smith2005development,clerkin2017real,zaadnoordijk2022lessonsnature,moulson2011neural,petroff2025world,franchak2024smith,skelton2022infant,schwarzer2014motor,johnson2010infantsmotion,myowa2011goal,arcaro2017monkeyface,wood2018development}. Second, an initially degraded (blurred, low-chroma) input may serve as an adaptive scaffold, biasing developmental trajectories toward useful inductive bias with lasting effects on perceptual organization \cite{vogelsang2024butterfly,vogelsang2018potential,vogelsang2024impact,vogelsang2025potential,mckyton2015limits}. 

Prior works in machine vision have leveraged some of these principles from developmental psychology. For example, some studies train models on longitudinal egocentric videos collected from head-mounted cameras worn by infants \cite{orhan2020self, sheybani2023curriculum, orhan2024learning}, but these works do not account for key characteristics of infant visual systems, such as low acuity and limited color sensitivity. More recent works have explored individual visual diets, such as progressive blur or color exposure \cite{vogelsang2024impact, vogelsang2025potential, vogelsang2018potential}, yet these models rely on fully supervised training with thousands of labeled examples, which is not ecologically plausible. 
To address these gaps, as shown in \textbf{Fig.~\ref{fig:illustration}}, we design developmentally inspired visual diets that capture key features of infant vision, following a progression from grayscale to color (\textbf{CDiet}), blur to sharp (\textbf{ADiet}), and preserving temporal continuity (\textbf{TDiet}), collectively termed \textbf{CATDiet}. We train self-supervised learning (SSL) models on object-centric videos under these constraints and examine how each diet and their combination shapes the emergence of robust visual representations. 

In parallel, SSL has recently emerged as a powerful approach for learning visual representations without labels. Prominent SSL methods \cite{chen2021simsam,bardes2021vicregbarlow,caron2020swav,henaff2020cpcv2,ermolov2021whiteningwmse} include SimCLR \cite{chen2020simple}, MoCo \cite{he2020momentummoco}, BYOL \cite{grill2020byol}, MAE \cite{he2022masked}, and DINO \cite{caron2021emerging}. However, these approaches largely rely on random data augmentations, leaving the potential of developmentally inspired visual diets underexplored. To investigate this, we introduce a comprehensive benchmark spanning 10 datasets for object recognition, texture–shape conflict tests, silhouette recognition, and depth perception, summarized in \textbf{Fig.~\ref{fig:illustration}}. On these benchmarks, CATDiet models, as well as their individual components, not only enhance robustness to visual corruptions but also exhibit developmental signatures consistent with biological vision, including changes in neural plasticity, the emergence of depth sensitivity, and behavioral patterns resembling infants in the visual cliff paradigm \cite{gibson1960viscliff}. 

To further leverage CATDiet, we introduce CombDiet, which starts training with CATDiet before transitioning to standard SSL while maintaining temporal continuity. This hybrid approach combines the ecological grounding of early visual experience with the efficiency of mature vision, consistently outperforming standard SSL on both in-domain and out-of-domain image recognition and depth perception tasks. Our main contributions are as follows:

\noindent\textbf{1.} We introduce CATDiet, a developmentally inspired visual diet in SSL that emulates the progression of infant vision through 3 staged constraints: grayscale-to-color (C), blur-to-sharp (A), and preserved temporal continuity (T).

\noindent\textbf{2.} We establish comprehensive benchmarks across 10 datasets covering object recognition and depth perception. The evaluation suite includes corrupted image recognition, texture–shape cue conflict tests, silhouette recognition, depth-order classification, and the visual cliff paradigm. This framework allows systematic, quantitative assessment of how individual visual diets and their combinations improve visual robustness.

\noindent\textbf{3.} Remarkably, even when trained exclusively on object-centric videos—without supervision from infant behavior or monkey neural data—CATDiet exhibits developmental signatures aligned with biological vision, including neural plasticity changes consistent with synaptic density in macaque V1, early emergent depth sensitivity, and behavioral patterns resembling infants’ responses in the visual cliff paradigm. 

\noindent\textbf{4.} Building on these insights, we introduce CombDiet, which initializes standard SSL with CATDiet while maintaining temporal continuity. CombDiet substantially outperforms standard SSL across our benchmarks, underscoring the practical benefits of early visual diets for real-world computer vision applications.

\begin{figure*}[t] 
    \centering
    \includegraphics[width=\textwidth]{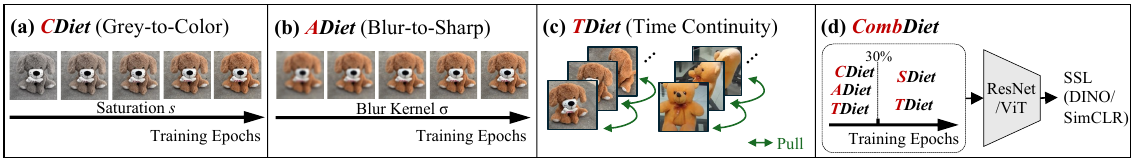}
    \vspace{-4mm}
    \caption{
    \textbf{Overview of our proposed developmental visual diets.}
\textit{CAT}Diet (panels a–c) integrates 3 individual visual diets detailed in \textbf{\cref{sec:CAT}}: 
(a) CDiet, in which image saturation gradually increases as chromatic information is progressively introduced throughout training; 
(b) ADiet, where the standard deviation $\sigma$ of Gaussian blur kernels decreases, enhancing spatial details over time; and 
(c) TDiet, which encourages representations of adjacent views of the same object to be closer, capturing temporal continuity in object-centric videos.  
(d) CombDiet extends CATDiet to a more general setting. In the first phase, CATDiet serves as a warm-up stage spanning the initial 30\% of training epochs. In the second phase, CombDiet transitions to the Standard Diet (SDiet) while retaining Temporality-Diet (TDiet). SDiet corresponds to the standard data augmentation pipeline used in conventional SSL training regimes. We evaluate CombDiet using 2 representative SSL methods (SimCLR \cite{chen2020simple} and DINO \cite{caron2021emerging}) with 2 widely adopted backbones: ResNet \cite{he2016deep} and ViT \cite{dosovitskiy2020image}.
}
\vspace{-4mm}
    
    \label{fig:method}
\end{figure*}

\section{Related Works}

\noindent\textbf{Developing AI models inspired by developmental psychology.}
A growing body of research draws inspiration from developmental psychology to inform AI model design. Existing approaches broadly follow two directions: (1) pretraining on children's egocentric videos to capture spatial statistics, multimodal associations, and infant-perspective object distributions \cite{sullivan2021saycam,orhan2020self,sheybani2023curriculum,orhan2024learning}, and (2) modeling specific developmental transitions—such as from blur to sharp or grayscale to color vision—under controlled, supervised settings \cite{ayzenberg2025fast,vogelsang2024impact,vogelsang2018potential,vogelsang2024butterfly,vogelsang2025potential,lu2025adoptingDVD}.
While these studies provide valuable insights, they have key limitations. Egocentric videos approximate viewing statistics but overlook crucial properties of infants’ visual systems, such as low spatial acuity and limited color sensitivity, which we explicitly model. Moreover, training directly on raw video streams makes it difficult to disentangle which factors contribute to generalizable object representations; our benchmark enables systematic, quantitative evaluation of individual and combined visual diets. Prior evaluations also emphasize clean-image recognition, neglecting model robustness and generalization, which we address through extensive out-of-domain testing. Finally, most prior works rely on supervised learning, reducing ecological validity and scalability; in contrast, our framework embeds developmental principles within a self-supervised paradigm to enable fully label-free, ecologically grounded learning.

\noindent\textbf{Self-Supervised Learning.}
Self-Supervised Learning (SSL) has become a dominant paradigm for learning visual representations without manual annotations. Existing SSL methods broadly fall into four categories: (1) contrastive methods \cite{chen2020improvedmocov2,chen2021empiricalmocov3,chen2020simclrv2,wu2018InstDisc,ye2019invaspread,henaff2020cpcv2}, such as SimCLR \cite{chen2020simple} and MoCo \cite{he2020momentummoco}, which align augmented views of the same image while separating different ones; (2) non-contrastive methods \cite{bardes2021vicregbarlow,koohpayegani2021meanbyol,ermolov2021whiteningwmse}, including BYOL \cite{grill2020byol}, SimSiam \cite{chen2021simsam}, and Barlow Twins \cite{zbontar2021barlow}, which achieve invariance without negative pairs through architectural or loss-based constraints that prevent representational collapse; (3) generative methods \cite{tong2022videomae,wang2023videomae2}, such as autoencoders \cite{hinton1993autoencoders} and MAE \cite{he2022masked}, which reconstruct missing or corrupted inputs; and (4) clustering-based methods \cite{kim2022selfdino,oquab2023dinov2,simeoni2025dinov3}, such as SwAV \cite{caron2020swav} and DINO \cite{caron2021emerging}, which align features with evolving cluster prototypes. While these methods rely on random data augmentations to enrich training signals, they overlook the developmental structure of visual experience. Building on video-based SSL frameworks that leverage temporal continuity across frames \cite{schneider2021simclrtt,oord2018cpcv1,qian2021spatiotemporalvideocl1,han2020videocl2,feichtenhofer2021videocl3,wang2024poodlevideocl4,aubret2024interactionvideocl5,parthasarathy2023inavideocl6,wu2021contrastivevideocl7,tschannen2020selfvideocl8,kuang2021videocl9}, we introduce TDiet, which explicitly enforces temporal alignment between adjacent frames to capture the natural continuity of visual experience. Unlike prior video-based SSL models, which are typically evaluated only on video understanding benchmarks, we also assess them on corrupted image recognition and depth perception tasks. Furthermore, by integrating TDiet with the CDiet and ADiet, our combined CATDiet framework achieves a synergistic performance improvement that surpasses the contribution of any individual component.

\noindent \textbf{Improving Model Robustness to Visual Corruptions.}
Out-of-distribution (OOD) generalization in visual representation learning remains a longstanding challenge \cite{madan2021smallrf3fool,joshi2019semanticrf10fool,zhang2019makingrf1,chaman2021trulyrf2,beery2018recognitionrf4,zhang2024adversarialrf5,barbu2019objectnetrf6,liu2018beyondrf7,zeng2019adversarialrf8,sakai2022threerf9}.
Existing approaches have explored a wide range of strategies, including specialized training objectives \cite{ren2019likelihoodrf30,sastry2019detectingrf31,hodge2004surveyrf32,aggarwal2001outlierrf33}, large-scale and diverse datasets \cite{oquab2023dinov2,simeoni2025dinov3}, advanced data augmentations \cite{hendrycks2021many,yun2019cutmixrf12,hendrycks2019augmixrf13,zhang2017mixuprf14}, domain-invariant feature learning \cite{muandet2013domainrf17,li2018domainrf18,li2018deeprf19,motiian2017unifiedrf20,shao2019multirf21,wang2021respectingrf22}, generative modeling \cite{ilse2020divarf15,wang2020rf16}, and architectural innovations \cite{shahtalebi2021sandrf23,sun2016deeprf24,arjovsky2019invariantrf25,kim2021selfregrf26,vedantam2021empiricalrf27,krueger2021outrf28,blanchard2021domainrf29}.
While these approaches improve robustness on benchmarks like ImageNet-C \cite{hendrycks2019benchmarking}, they often depend on extensive artificial data augmentations or non-ecological training objectives, instead of fostering emergent generalization from limited and naturalistic visual experience as seen in humans \cite{mintun2021interaction}.
In contrast, we examine how early staged visual diets, inspired by human visual development and learned via self-supervision, can enhance model robustness across diverse in-domain and out-of-domain benchmarks spanning 10 datasets.

\section{Our Proposed Developmental Visual Diets}
We propose CombDiet (\textbf{\cref{fig:method}d}), a two-phase self-supervised learning (SSL) framework that embeds principles of human visual development into both the data curriculum and the learning objectives. In the first phase, CATDiet serves as a warm-up stage that mirrors infants’ first-year visual progression, marked by rapid perceptual development toward near-mature vision. This phase spans the initial 30\% of training epochs and integrates two data curricula, Color-Diet (CDiet) and Acuity-Diet (ADiet), with a temporal regularization objective, Temporality-Diet (TDiet), to emulate the temporal continuity inherent in early visual experience. In the second phase, Standard Diet (SDiet) represents mature visual experience. CombDiet transitions to SDiet while retaining TDiet, thereby maintaining temporal coherence throughout learning.

\subsection{Phase 1 - CATDiet} \label{sec:CAT}

\noindent \textbf{Color-Diet (CDiet).}  
To mimic the grey-to-color progression, we design a five-stage saturation schedule (\textbf{\cref{fig:method}a}), where each stage specifies a blend ratio $s$ between the fully colored image $I_c$ and its grayscale counterpart $I_g$. From the first to fifth stages, $s$ is sampled within (0.20, 0.36), (0.36, 0.52), (0.52, 0.68), (0.68, 0.84), and (0.84, 1.0), respectively.  
For example, in the first stage, for any given training image, a random $s$ is sampled within (0.2, 0.36) and used to blend its grayscale and colored versions:
$s I_c + (1-s) I_g$.
Stage durations $[10,7,6,5,2]$, measured in training epochs, decrease across stages. This design choice reflects the rapid increase in chromatic sensitivity during early infancy \cite{dobkins2001colorspeed,crognale2002development, vogelsang2024impact, maule2023development, skelton2022infant, werner1979human}.

\noindent \textbf{Acuity-Diet (ADiet).}  
For the blur-to-sharp curriculum, we define a five-stage Gaussian blur schedule (\textbf{\cref{fig:method}b}). Across stages one to five, each training image is blurred with standard deviations $\sigma \in [4, 3, 2, 1, 0]$, corresponding to kernel sizes $[25, 19, 13, 7, 1]$ pixels for images of size 224×224. Stage durations $[10, 6, 6, 3, 5]$, measured in training epochs, decrease across stages. This design reflects developmental psychology findings that visual acuity increases approximately exponentially during the first year of life \cite{norcia1985spatial, teller1997first}. In CATDiet, the ADiet and CDiet are integrated by interleaving the Gaussian blur schedule of ADiet with the saturation schedule of CDiet, with each schedule maintaining its own stage durations.

\noindent \textbf{Temporality-Diet (TDiet).} 
We introduce a temporal alignment objective (\textbf{\cref{fig:method}c}) to capture temporal continuity. Adjacent video frames are temporally and spatially related, often depicting the same object undergoing small, continuous changes in viewpoint. This temporal coherence provides an intrinsic, free supervisory signal, encouraging SSL models to learn view-invariant representations by pulling embeddings of adjacent frames closer \cite{wood2018development}. Combined with CDiet and ADiet, which apply augmentations to each frame, the augmented versions of adjacent frames are grouped as a positive set. The training objective is therefore to bring together the representations of adjacent frames and their augmented variants.

\subsection{Phase 2 - SDiet and TDiet}
CombDiet allocates the first 30\% of training epochs to Phase 1 and the remaining 70\% to Phase 2, with these hyperparameters determined via grid search to optimize performance. In Phase 1, CATDiet emulates the progression of early visual diets during the first year of life, while Phase 2 transitions to SDiet to approximate adult visual experience. In \textbf{SDiet}, standard data augmentation techniques from the 2 representative SSL methods described below are applied to training images. The TDiet objective remains unchanged in Phase 2, consistently encouraging alignment of representations between adjacent frames and their augmented versions throughout training.

\noindent\textbf{SSL Methods and Model Backbones.} 
Ideally, our visual diets could be applied to all SSL methods and model backbones, but exhaustively evaluating all combinations is computationally prohibitive. We therefore select 2 representative SSL methods, SimCLR \cite{chen2020simple} and DINO \cite{caron2021emerging}. Unlike generative SSL methods such as MAE \cite{halvagal2023combination}, which reconstruct missing or corrupted pixels and are less ecologically valid, SimCLR and DINO are widely used and yield more biologically plausible SSL representations \cite{halvagal2023combination,yamamoto2025dinoplausible,yerxa2024contrastiveplausible,zhuang2021simclrplausible1,konkle2022simclrplausible2,parthasarathy2024simclrplausible3}.
Given their distinct training objectives, we adapt TDiet to each method. In SimCLR, TDiet pulls adjacent frames and their augmented views closer in embedding space while keeping the original negative pairs to push apart representations of non-adjacent frames within the same video or frames from different videos. In DINO, TDiet aligns features of adjacent frames and their augmented views through evolving cluster prototypes from the student and teacher networks.
For each SSL method, we experiment with 2 common backbones: the 2D-CNN ResNet \cite{he2016deep} and the transformer-based ViT \cite{dosovitskiy2020image}, yielding 4 model variants denoted as [SSL method]-[backbone] (e.g., SimCLR-ResNet).

\noindent\textbf{Implementation Details.} 
All models are trained with a batch size of 64 and an image resolution of $224 \times 224$. We use the AdamW optimizer \cite{loshchilov2017adamw} with a learning rate of $5\times10^{-4}$, weight decay of $1\times10^{-4}$, and a cosine annealing schedule with a 10-epoch warm-up. Experiments are conducted on NVIDIA RTX A6000 and RTX 6000 Ada Generation GPUs. Additional implementation details, and parameter analyses are provided in \textbf{\cref{sec:temp}}. Additional analysis on variations in training schedules across different diets, as well as the design of TDiet, are provided in \textbf{\cref{sec:suppablation}}.

\section{Benchmarking Developmental Visual Diets}
In this section, we present a comprehensive benchmark for evaluating developmental visual diets. It encompasses 10 datasets, multiple evaluation metrics across diverse downstream tasks, and baseline models for comparison with our proposed developmental visual diets.

\subsection{Datasets} \label{sec:datasets}
\noindent \textbf{CO3D} \cite{reizenstein21co3d}:
The CO3D dataset comprises real-world, object-centric videos capturing 360° azimuthal rotations around individual objects. We select 10 object categories with about 4,500 instances in total. Training and test splits are defined at the instance level within each category. During training, we uniformly sample 10 frames per video to span the complete viewing circle, corresponding to roughly 36° between adjacent views. For testing, we sample 2 frames per video at the same interval as training for diverse viewpoints with minimal redundancy.

\noindent \textbf{CO3D-C}:
To evaluate model robustness in object recognition, we construct a corrupted variant of the CO3D test set, using the distortion operators defined in the ImageNet-C protocol \cite{hendrycks2019benchmarking}. CO3D-C comprises 15 corruption types across 4 families—noise, blur, weather, and digital artifacts—each applied at 5 severity levels.

\noindent\textbf{3D-PC} \cite{linsley20243d}: The 3D-PC Depth Order dataset contains ~4,500 images of rendered 3D scenes. Each image depicts a green arrow (camera viewpoint) and a red ball. The task is formulated as a binary classification problem, where the model predicts whether the arrow is closer than the ball (\textbf{\cref{fig:illustration}e}). We adopt the official training and test splits.

\noindent\textbf{IEVC}: The classical visual cliff paradigm in developmental psychology places infants on a glass platform with one side appearing shallow (texture directly beneath the glass) and the other deep (texture several centimeters below). Infants’ hesitation to cross the “cliff” marks the emergence of depth perception. To simulate this, we render 3 infant-perspective views in Blender facing the deep side (\textbf{\cref{fig:illustration}f} and \textbf{\cref{fig:behavior}c}).
Similar to 3D-PC, we formulate this task as a binary depth-order classification task for the models.


\noindent\textbf{SAYCam (SAY)} \cite{sullivan2021saycam}: SAY is a longitudinal egocentric video dataset capturing the daily visual experiences of 3 infants (S, A, and Y) from 6 to 32 months of age, providing spatial statistics, infant-perspective object distributions, and temporal continuity reflective of early visual development. In our experiments, we use recordings from child S, which include an annotated subset of video frames across 26 object categories. The unannotated portion is used for self-supervised pretraining, while the annotated frames serve as the linear probing dataset. From each video, we sample a 2-second clip every 2 minutes to minimize redundancy, with each clip sampled at 5 fps, yielding approximately 35,000 frames for pretraining.

\noindent\textbf{SAYCam-C (SAY-C)}: A corrupted extension of SAY is constructed using the same 4 corruption families as the CO3D-C dataset. This dataset evaluates model robustness in object recognition under corrupted egocentric views.

\noindent\textbf{ImageNet-21K (IN)} \cite{ridnik2021in21k}: IN is a large-scale naturalistic image dataset containing over 14 million images across 21,000 object categories. For our experiments, we select 
15 overlapping classes between IN and SAY. 
The list of overlapping classes is provided in \textbf{\cref{sec:temp}}.
\noindent\textbf{ImageNet-C (IN-C)}: Corrupted IN is constructed using the same 4 corruption families as CO3D-C.

\noindent\textbf{Texture–Shape Cue Conflict (TSCC)} \cite{geirhos2018imagenet}: TSCC is a diagnostic dataset where each image fuses the global shape of one object category with the local texture of another via style transfer (\textbf{\cref{fig:illustration}c}). It evaluates whether SSL models rely primarily on shape or texture for object recognition. We select 3 overlapping object classes between TSCC and SAY for evaluation.

\noindent\textbf{Silhouettes-Only (Sil-O)} \cite{geirhos2018imagenet}: Sil-O is a texture-free variant of IN where each object is rendered as a black silhouette on a white background (\textbf{\cref{fig:illustration}d}). We select the same 3 overlapping classes as in TSCC. This dataset assesses shape-based recognition independent of color.

\subsection{Baselines}

\noindent \textbf{Baselines for CDiet and ADiet.}
We define 4 baselines by altering the order of their staged curricula. For brevity, we denote each baseline together with its corresponding visual diet as \textbf{[Diet]-[Baseline]}; for example, \textbf{C-REV} indicates the REV baseline applied to CDiet. \textbf{Reverse Order (REV).} The progression direction and the allocation of stage durations in the staged schedule are both reversed; for instance, instead of progressing from grayscale to color in CDiet, \textbf{C-REV} applies the five-stage saturation schedule in reverse, from color to grayscale. \textbf{Shuffled Order (SHF).} The original stage ordering is discarded, and data from all 5 stages are merged into a single pool, with each mini-batch randomly sampled from this pool during SSL pretraining. \textbf{First-Only (FO).} To limit SSL models to early-stage inputs, they are pretrained only on the first stage of each visual diet; for example, in \textbf{C-FO}, only the lowest blend ratio $s$ from stage 1 is applied across all 5 stages in CDiet.
\textbf{Last-Only (LO).} LO uses all the original images without any color or blur modifications. For example, \textbf{C-LO} uses the original full-color images for pretraining.



\noindent \textbf{Baselines for TDiet.}
We include a single baseline, \textbf{Non-Smooth}, in which the objective pulls together only cropped views from the same frame, without encouraging similarity between neighboring frames. 

\noindent \textbf{Baselines for CATDiet.}
We adopt the same 4 baselines as for CDiet and ADiet, applying each baseline to its respective schedule individually and then combining them to form the corresponding baselines, such as \textbf{CAT-REV}. 

\noindent \textbf{Baselines for CombDiet.}
We compare \textbf{CombDiet} against two controlled baselines:
\noindent \textbf{Shuffled Order (SHF).}
This variant follows the same two-phase schedule as CombDiet, with CAT-SHF used in Phase 1 followed by SDiet in Phase 2. It isolates the effect of the developmental order while maintaining the same data distribution, architecture, and training duration as CombDiet.
\noindent \textbf{Standard SSL (STD).}
We include standard SSL baselines, SimCLR and DINO. For fair comparisons, we use the same backbones as in CombDiet but retain their conventional training paradigm with standard data augmentations and loss objectives.

\subsection{Evaluation Metrics} \label{sec:evalmetrics}

\noindent \textbf{Metrics for object recognition.} \textbf{Top-1 accuracy (Acc)} denotes top-1 classification accuracy. The \textbf{mean Corruption Error (mCE)} \cite{hendrycks2019benchmarking} is computed as the average normalized error across all corruption types and severity levels, with lower values indicating stronger model robustness. See \textbf{\cref{sec:temp}} for details.
\textbf{Shape-Bias (S-Bias)} quantifies perceptual alignment with shape cues in the TSCC dataset. It is defined as the proportion of shape-consistent predictions among all images classified as either shape- or texture-consistent by any of the three models (CombDiet, SHF, STD). Higher S-Bias indicates stronger reliance on global shape cues over local textures.


\noindent \textbf{Metrics for quantifying network connectivity.} \textbf{Fisher Information Matrix (FIM)} is defined as the trace of the Fisher Information Matrix \cite{achille2018critical}, computed from the gradients of the network parameters. Intuitively, it reflects the network’s sensitivity to small parameter perturbations and its adaptability during SSL pretraining.


\noindent \textbf{Metrics for depth-order classification.} \textbf{Depth Accuracy (dAcc)} denotes the binary classification accuracy on the depth-order prediction task. dAcc of 0.5 corresponds to chance performance, reflecting random guesses on whether the green arrow is closer than the red ball.

\section{Results} 
\subsection{Evaluation of CATDiet on Object Recognition}
\label{sec:ablation}
We pretrain 4 SSL models on the integrated CATDiet as well as its individual diets using the CO3D training set, followed by training 10-way linear probes on the same set. The models and their baselines are then evaluated on clean CO3D test images and corrupted CO3D-C images. Results for SimCLR-ResNet are shown in the main text, with other models presented in \textbf{\cref{sec:otherablation}}. Consistent observations and analyses apply across all models. 

\begin{figure}[t]
  \centering
  \includegraphics[width=0.9\columnwidth]{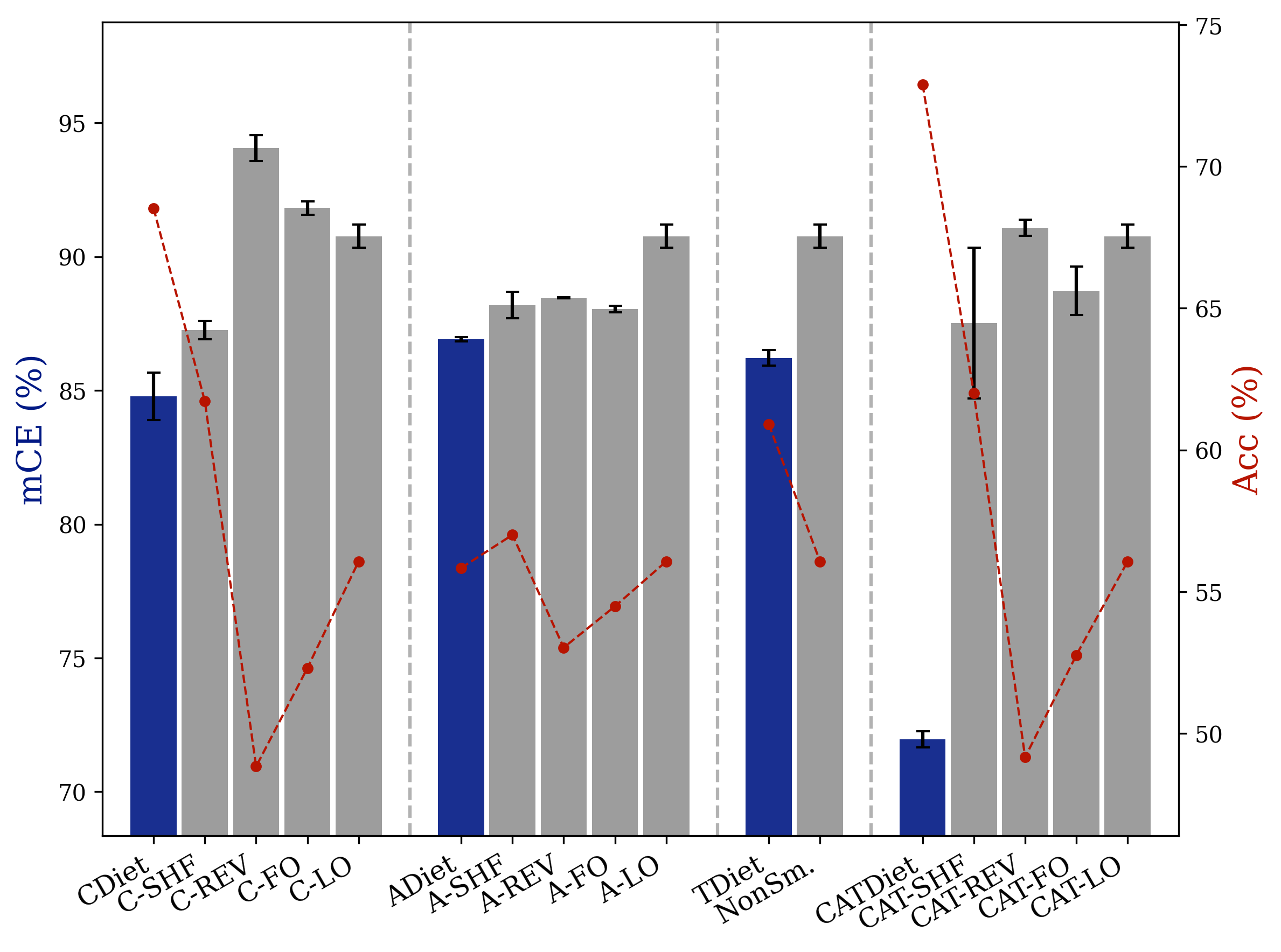}
  \vspace{-4mm}
  \caption{\textbf{Object recognition performance on CO3D (clean) and CO3D-C (corrupted) datasets for SimCLR-ResNet pretrained on CATDiet and its individual diets.} Bars show mCE ($\downarrow$, left axis), where blue and gray bars denote our proposed visual diets and their corresponding baselines; the dashed red line indicates Acc ($\uparrow$, right axis). Error bars represent the standard error of the mean (SEM) of mCE over three runs. The four panels correspond to different attributes of the proposed visual diets: Color, Acuity, Temporality, and their combination (see \textbf{\cref{sec:ablation}}). 
}
\vspace{-4mm}
  \label{fig:simclr-resent}
\end{figure}

\begin{figure*}[t]
  \centering
  \includegraphics[width=0.9\textwidth]{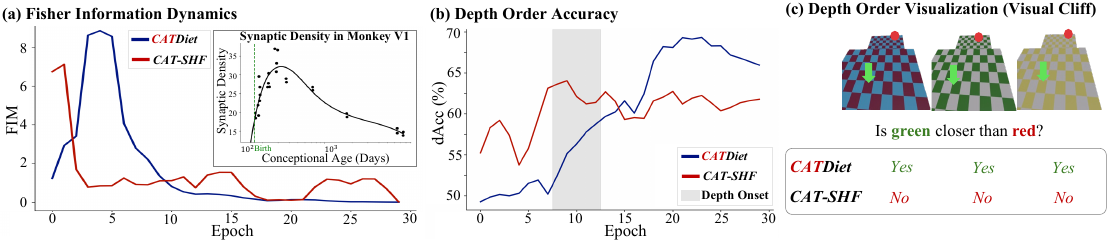}
  \vspace{-4mm}
  \caption{
  \textbf{Signature developmental patterns observed in SimCLR-ResNet pretrained on CATDiet from CO3D (a), 3D-PC (b), and IEVC (c).}  
(a) The trace of the Fisher Information Matrix (FIM) for the SSL model gradients \cite{achille2018critical} is plotted across pretraining epochs, capturing the sensitivity of network outputs to small weight perturbations. The inset shows synaptic density changes in macaque primary visual cortex (V1) \cite{rakic1986concurrent}, highlighting a similar rise-and-fall pattern for the model pretrained on CATDiet (blue) compared to FIM changes in CAT-SHF (red).
(b) Binary classification accuracy on a depth-order task as a function of pretraining epochs. Blue and red curves correspond to CATDiet and SHF, respectively. The shaded region marks the period of rapid accuracy increase in dAcc for CATDiet.  
(c) Simulated Visual Cliff experiment. The top row shows egocentric views from an infant’s perspective crawling on a glass platform (see \textbf{\cref{sec:datasets}}). The table below summarizes model responses for CATDiet and CAT-SHF to the binary question ``Is the green arrow closer than the red ball?'' (``yes'' indicates that the green arrow is closer).
} 
\vspace{-4mm}
  \label{fig:behavior}
\end{figure*}

\noindent \textbf{Natural progression order matters and incorrect progression order impairs model robustness.} From the first two panels of \textbf{Fig.~\ref{fig:simclr-resent}}, both CDiet and ADiet models pretrained with the SHF or REV exhibit higher mCE compared to their original staged curricula. For example, CDiet outperforms the best-performing baseline \textbf{C-SHF} by 2.5\% in mCE. This indicates that maintaining the original schedule in visual diets is critical for learning robust object representations, as it provides the correct inductive bias. 
Interestingly, models pretrained with REV may even underperform SHF (e.g., \textbf{C-REV} with mCE of 94.1\% versus \textbf{C-SHF} with mCE of 87.3\%), suggesting that incorrect progression orders can sometimes disrupt the model’s ability to capture meaningful visual features.

\noindent \textbf{Early limitations in visual diets are beneficial for developmental scaffolding but insufficient on their own.} To probe the effect of early-stage and late-stage diets on model robustness, we compare the FO and LO baselines against our proposed CDiet and ADiet (first two panels of \textbf{Fig.~\ref{fig:simclr-resent}}). Pretraining on LO is inferior to the full staged curriculum, suggesting that limited early-stage exposure helps scaffold representation learning. For example, CDiet achieves an mCE of 84.8\%, compared to 90.8\% for C-LO. However, FO alone does not match the performance of our diet (e.g., CDiet vs. C-FO and ADiet vs. A-FO), demonstrating that early constraints facilitate learning but require subsequent stages to fully develop robust representations. These results echo two observations from developmental psychology. First, children who begin visual experience with relatively high acuity due to early cataract removal can discriminate faces based on local features but fail to detect their configural changes \cite{maurer2007sleeper,vogelsang2018potential}. Second, children whose vision begins with mature cone cells due to early cataract removal show a marked decrease in recognizing grayscale images relative to color images \cite{vogelsang2024impact}.

\noindent \textbf{Models achieve strong accuracy on clean images while exhibiting robustness to image corruptions.} From the first two panels of \textbf{Fig.~\ref{fig:simclr-resent}}, across CDiet and ADiet, models pretrained with our proposed diets not only perform competitively well with or better than all baselines (REV, SHF, FO, LO) in terms of Acc on clean images, but also significantly outperform all baselines in terms of mCE on corrupted images. This demonstrates that developmental-inspired training simultaneously enhances object recognition accuracy and model robustness. 

\noindent \textbf{Temporal continuity provides free and useful regularization to learn robust object representations.} In TDiet (third panel of \textbf{Fig.~\ref{fig:simclr-resent}}), including the training objective of temporal continuity leads to consistent gains in both Acc and mCE compared to the Non-Smooth baseline. By encouraging similarity between neighboring frames, TDiet effectively regularizes the learned representations, improving robustness to natural image corruptions.

\noindent \textbf{Integrated CATDiet achieves more than its individual components.} In the fourth panel of \textbf{Fig.~\ref{fig:simclr-resent}}, CATDiet achieves a substantially lower mCE of 72.0\% and a higher Acc of 72.9\% compared to its individual components (CDiet: 84.8\% in mCE and 68.5\% in Acc, ADiet: 86.9\% in mCE and 55.8\% in Acc, TDiet: 86.2\% in mCE and 60.9\% in Acc). This demonstrates that the combined curriculum provides synergistic benefits beyond the sum of its parts.  
Furthermore, we compare CATDiet against its baselines in both mCE and Acc. Consistent with observations in CDiet and ADiet individually, the baselines underperform CATDiet, with performance gaps even larger than those observed for the individual diets, further highlighting the importance of integrating developmentally inspired diets.

\noindent \textbf{Changes in network connectivity of SSL models pretrained on CATDiet mirror synaptic density changes in macaque V1 over development.}  
As shown in \textbf{\cref{fig:behavior}a}, network connectivity measured via FIM for CATDiet closely follows synaptic density changes in macaque primary visual cortex across the lifespan \cite{rakic1986concurrent}. FIM rises sharply during early pretraining, peaks around the 5th epoch, and gradually declines, indicating an initial phase of high plasticity followed by progressive weight consolidation. This suggests that CATDiet undergoes an early exploratory stage of representation formation before stabilizing into robust feature embeddings. In contrast, CAT-SHF shows a monotonically decreasing curve, implying premature convergence. These results indicate that a structured developmental diet balances exploration and exploitation, with the model first exploring the embedding space and later consolidating robust representations through weight pruning.

\subsection{Evaluation of CATDiet on Depth Perception}
\label{sec:depth}
We pretrain the SimCLR-ResNet on CATDiet using the CO3D training set, followed by training its linear probes on the 3D-PC training set for binary depth-order classification. The model and its baselines are then evaluated on the 3D-PC test set and the IEVC dataset, which simulates the Visual Cliff paradigm \cite{gibson1960viscliff}. 

\noindent \textbf{Progressive depth perception emerges during pretraining with infant-like staged diets.}  
As shown in \textbf{\cref{fig:behavior}b}, the dAcc on the 3D-PC dataset for the SSL model pretrained on CATDiet increases sharply around the fifth epoch, marking the emergence of depth sensitivity. In contrast, CAT-SHF plateaus early with a lower dAcc of 61.5\%, indicating that disrupting the developmental sequence impairs acquisition of depth cues. This result suggests that pictorial depth sensitivity in infants emerges postnatally rather than being innate \cite{aslin2001visual}.

\noindent \textbf{Models pretrained with progressive staged diets align with infant behaviors in the visual cliff paradigm.}  
In the visual cliff evaluation (\textbf{\cref{fig:behavior}c}), the SSL model pretrained on CATDiet correctly predicts the depth order across all three synthesized environments in the IEVC dataset. Specifically, the model identifies the shallow side as safe and avoids the deep side, mirroring the avoidance behavior observed in one-year-old infants \cite{gibson1960viscliff}. In contrast, CAT-SHF fails to make accurate depth-order predictions, underscoring the importance of developmental order in visual diets. Consistent with \cite{aslin2001visual}, these results also suggest that depth perception is not innate but rather emerges through early visual experience.

\begin{table}[t]
\resizebox{\columnwidth}{!}{
\large
\begin{tabular}{cllccccccc}
\toprule
\multicolumn{1}{l}{\textbf{}}           & \textbf{}          & \textbf{}                                                                 & \textbf{SAY}                          & \textbf{SAY-C}                        & \textbf{IN}                           & \textbf{IN-C}                         & \textbf{TSCC}                         & \textbf{Sil-O}                        & \textbf{3D-PC}                        \\
\multicolumn{1}{l}{\textbf{}}           & \textbf{}          & \textbf{}                                                                 & \textbf{\cite{sullivan2021saycam}}                            & \textbf{\cite{hendrycks2019benchmarking}}                            & \textbf{\cite{ridnik2021in21k}}                            & \textbf{\cite{hendrycks2019benchmarking}}                            & \textbf{\cite{geirhos2018imagenet}}                            & \textbf{\cite{geirhos2018imagenet}}                            & \textbf{\cite{linsley20243d}}                            \\
\multicolumn{1}{l}{\textbf{}}           & \textbf{}          & \textbf{}                                                                 & \textbf{Acc}                         & \textbf{mCE}                          & \textbf{Acc}                         & \textbf{mCE}                          & \textbf{S-Bias}                       & \textbf{Acc}                          & \textbf{dAcc}                         \\ \midrule
\multicolumn{2}{c}{} & \cellcolor{gray!15}\textbf{Comb} 
& \cellcolor{gray!15}\textbf{54.5} \small $\pm$ 0.3 
& \cellcolor{gray!15}\textbf{79.0} \small $\pm$ 0.1
& \cellcolor{gray!15}\textbf{63.7} \small $\pm$ 0.3 
& \cellcolor{gray!15}\textbf{71.5} \small $\pm$ 0.2 
& \cellcolor{gray!15}\textbf{29.1} \small $\pm$ 3.5 
& \cellcolor{gray!15}\textbf{41.1} \small $\pm$ 4.0 
& \cellcolor{gray!15}\textbf{72.5} \small $\pm$ 1.0  \\
\multicolumn{2}{c}{} & \textbf{SHF} 
& 50.0 \small $\pm$ 0.2
& 85.5 \small $\pm$ 0.2
& 62.0 \small $\pm$ 0.6
& 80.2 \small $\pm$ 0.3
& 21.1 \small $\pm$ 4.6
& 34.4 \small $\pm$ 1.1
& 66.6 \small $\pm$ 1.2    \\
\multicolumn{2}{c}{\multirow{-3}{*}{\textbf{\makecell{S-V\\\cite{chen2020simple}\cite{dosovitskiy2020image}}}}}    & \textbf{STD}                                       
& 48.8 \small $\pm$ 0.1
& 86.9 \small $\pm$ 0.1
& 63.2 \small $\pm$ 0.4
& 77.0 \small $\pm$ 0.3
& 19.8 \small $\pm$ 0.8
& \textbf{41.1} \small $\pm$ 2.9
& 64.4 \small $\pm$ 1.8  \\ \midrule
\multicolumn{2}{c}{} & \cellcolor{gray!15}\textbf{Comb} 
& \cellcolor{gray!15}\textbf{63.0} \small $\pm$ 0.1
& \cellcolor{gray!15}\textbf{77.1} \small $\pm$ 0.1
& \cellcolor{gray!15}74.6 \small $\pm$ 0.7
& \cellcolor{gray!15}\textbf{64.8} \small $\pm$ 0.6
& \cellcolor{gray!15}\textbf{45.1} \small $\pm$ 2.8
& \cellcolor{gray!15}41.1 \small $\pm$ 2.9
& \cellcolor{gray!15}\textbf{68.6} \small $\pm$ 1.0 \\
\multicolumn{2}{c}{} & \textbf{SHF} 
& 56.3 \small $\pm$ 0.3
& 86.4 \small $\pm$ 0.2
& 72.4 \small $\pm$ 0.3
& 73.2 \small $\pm$ 0.8
& 34.2 \small $\pm$ 2.1
& 51.1 \small $\pm$ 4.8
& 61.8 \small $\pm$ 0.8  \\
\multicolumn{2}{c}{\multirow{-3}{*}{\textbf{\makecell{S-R\\\cite{chen2020simple}\cite{he2016deep}}}}} & \textbf{STD}        
& 54.9 \small $\pm$ 0.1
& 85.7 \small $\pm$ 0.1
& \textbf{75.2} \small $\pm$ 0.9
& 71.3 \small $\pm$ 0.1
& 41.1 \small $\pm$ 1.6
& \textbf{54.4} \small $\pm$ 1.1
& 63.9 \small $\pm$ 0.3           \\ \midrule
\multicolumn{2}{c}{} & \cellcolor{gray!15}\textbf{Comb} 
& \cellcolor{gray!15}\textbf{55.1} \small $\pm$ 0.3
& \cellcolor{gray!15}\textbf{76.0} \small $\pm$ 0.4
& \cellcolor{gray!15}66.1 \small $\pm$ 0.6
& \cellcolor{gray!15}70.3 \small $\pm$ 0.2
& \cellcolor{gray!15}\textbf{19.2} \small $\pm$ 2.6
& \cellcolor{gray!15}\textbf{62.2} \small $\pm$ 1.1
& \cellcolor{gray!15}\textbf{78.9} \small $\pm$ 0.9 \\
\multicolumn{2}{c}{} & \textbf{SHF} 
& 52.4 \small $\pm$ 0.1
& 81.0 \small $\pm$ 0.1
& 62.6 \small $\pm$ 0.3
& 77.5 \small $\pm$ 0.3
& 10.1 \small $\pm$ 0.1
& 33.3 \small $\pm$ 0.0
& 75.8 \small $\pm$ 0.9 \\
\multicolumn{2}{c}{\multirow{-3}{*}{\textbf{\makecell{D-V\\\cite{caron2021emerging}\cite{dosovitskiy2020image}}}}}      & \textbf{STD}                            
& 54.6 \small $\pm$ 0.2
& 79.6 \small $\pm$ 0.4
& \textbf{68.0} \small $\pm$ 0.4
& \textbf{70.2} \small $\pm$ 0.4
& 19.1 \small $\pm$ 1.0
& 51.1 \small $\pm$ 2.9
& 74.5 \small $\pm$ 0.6   \\ \midrule
\multicolumn{2}{c}{} & \cellcolor{gray!15}\textbf{Comb} 
& \cellcolor{gray!15}\textbf{59.9} \small $\pm$ 1.0
& \cellcolor{gray!15}\textbf{80.4} \small $\pm$ 0.3
& \cellcolor{gray!15}\textbf{75.9} \small $\pm$ 0.7
& \cellcolor{gray!15}\textbf{66.7} \small $\pm$ 0.4
& \cellcolor{gray!15}\textbf{24.0} \small $\pm$ 2.3
& \cellcolor{gray!15}\textbf{46.7} \small $\pm$ 3.9
& \cellcolor{gray!15}\textbf{66.9} \small $\pm$ 0.6\\
\multicolumn{2}{c}{} & \textbf{SHF} 
& 43.5 \small $\pm$ 0.3
& 100.3 \small $\pm$ 0.4
& 65.6 \small $\pm$ 0.6
& 89.6 \small $\pm$ 0.8
& 21.7 \small $\pm$ 0.4
& 40.0 \small $\pm$ 3.9
& 60.8 \small $\pm$ 3.2  \\
\multicolumn{2}{c}{\multirow{-3}{*}{\textbf{\makecell{D-R\\\cite{caron2021emerging}\cite{he2016deep}}}}}   & \textbf{STD}   
& 52.6 \small $\pm$ 0.5
& 91.8 \small $\pm$ 0.3
& 74.9 \small $\pm$ 0.4
& 77.4 \small $\pm$ 0.6
& 19.3 \small $\pm$ 1.2
& \textbf{46.7} \small $\pm$ 1.9
& 64.1 \small $\pm$ 0.5 \\ \bottomrule
\end{tabular}
}
\vspace{-2mm}
\caption{\textbf{Performance of SSL models pretrained on CombDiet from the SAY dataset and their baselines on object recognition and depth perception tasks.} Each row corresponds to a specific [SSL]-[backbone] configuration (four in total). In the first column, \textbf{S}, \textbf{V}, \textbf{R}, and \textbf{D} denote SimCLR \cite{chen2020simple}, ViT \cite{dosovitskiy2020image}, ResNet \cite{he2016deep}, and DINO \cite{caron2021emerging}, respectively; \textbf{Comb} indicates our proposed CombDiet. Within each row, three models are compared: CombDiet, SHF, and STD. Columns report results on the respective datasets with their corresponding evaluation metrics. Shaded rows highlight CombDiet performance. Values are reported as mean $\pm$ standard error of the mean (SEM) across three runs; best results are shown in \textbf{bold}.
}
\vspace{-6mm}
\label{tab:co3d-SAYCam}
\end{table}

\subsection{CombDiet Generalizes to Real-World Tasks} \label{sec:combdietsay}

We first pretrain four SSL models on CombDiet using the SAY training set followed by their linear probes trained on the annotated set for 26-way image classification. The models and their baselines are evaluated on the clean SAY test images and the corrupted SAY-C images for object recognition. Additionally, we train another set of linear probes on IN for 15-way image classification and evaluate them on out-of-domain object recognition datasets, including IN, IN-C, TSCC, and Sil-O. Finally, we train and evaluate their models with separate 2-way linear probes for the binary depth-order classification task on 3D-PC. We provide additional results for CO3D-pretrained SSL models in \textbf{\cref{sec:combdietco3d}}. Similar analyses apply to these models as well.

\noindent \textbf{CombDiet models achieve the best performance among all the baselines in object recognition and depth-order estimation tasks.}
As shown in Column 1 and Column 7 of \textbf{\cref{tab:co3d-SAYCam}}, models pretrained with CombDiet consistently outperform all baselines across the object recognition and depth-order classification benchmarks on the SAY and 3D-PC datasets.
For instance, for SimCLR-ViT, the CombDiet model achieves the highest Acc of 54.5\% on the clean SAY test images and the highest dAcc of 72.5\% on the 3D-PC dataset.
Interestingly, CombDiet-SHF performs comparably to STD, though both remain inferior to our CombDiet models. This indicates that the developmental order in the progressive schedule is critical; disrupting it degrades model performance to the level of STD. Additional analyses of our visual diets—including training convergence, comparisons with curriculum learning methods, the effectiveness on synthetic object manipulation videos, and statistical evaluations—are provided in \textbf{\cref{sec:suppanalysis}}.

\noindent \textbf{CombDiet models are more robust to recognizing out-of-domain images and exhibit stronger inductive biases towards shapes.}
Columns 2–6 in \textbf{\cref{tab:co3d-SAYCam}} show model performance on out-of-domain image recognition tasks. CombDiet models are more robust to corrupted images in SAY-C compared to all baselines. Remarkably, they also generalize to IN and IN-C—datasets never seen during training, which demonstrates strong out-of-distribution generalization capabilities. Despite never being exposed to TSCC images or any human behavioral data, CombDiet models exhibit human-like shape biases, which become even more pronounced on Sil-O, where object recognition relies solely on silhouettes. These results indicate that our developmental visual diets induce strong shape-based inductive biases, improving model generalization and practical applicability in real-world vision tasks.

\section{Discussion}
We present CATDiet, a progressive visual diet for SSL that simulates infant vision through staged constraints on color, acuity, and temporal continuity. Across ten comprehensive benchmarks, CATDiet enhances robustness in object recognition and exhibits biologically aligned developmental patterns, including neural plasticity changes, early emergence of depth perception, and infant-like hesitation in the visual cliff paradigm. Building on these insights, CombDiet integrates these developmental principles with standard SSL training, achieving superior in-domain and out-of-domain performance on object recognition and depth-order classification tasks. Our work provides an initial step toward modeling the progressive structure of early visual experience and offers a framework for scaffolding robust visual representations in machines. 
Further discussions on our work limitations and future works are presented in \textbf{\cref{sec:suppdiscussion}}.

\section*{Acknowledgement}
This research is supported by the National Research Foundation, Singapore under its NRFF award NRF-NRFF15-2023-0001 and Mengmi Zhang's Startup Grant from Nanyang Technological University, Singapore. We also gratefully acknowledge Databrary at New York University for granting us access to the SAYCam dataset.

{
    \small
    \bibliographystyle{ieeenat_fullname}
    \bibliography{main}

@String(ECCV= {Eur. Conf. Comput. Vis.})

@String(ICPR = {Int. Conf. Pattern Recog.})

@String(ICIP = {IEEE Int. Conf. Image Process.})

@String(AAAI = {AAAI})

@String(ECCV  = {ECCV})

@String(ICPR  = {ICPR})

@String(ICIP  = {ICIP})

@inproceedings{zhang2017deep,
  title={Deep future gaze: Gaze anticipation on egocentric videos using adversarial networks},
  author={Zhang, Mengmi and Teck Ma, Keng and Hwee Lim, Joo and Zhao, Qi and Feng, Jiashi},
  booktitle={Proceedings of the IEEE conference on computer vision and pattern recognition},
  pages={4372--4381},
  year={2017}
}

@article{zhang2018anticipating,
  title={Anticipating where people will look using adversarial networks},
  author={Zhang, Mengmi and Ma, Keng Teck and Lim, Joo Hwee and Zhao, Qi and Feng, Jiashi},
  journal={IEEE transactions on pattern analysis and machine intelligence},
  volume={41},
  number={8},
  pages={1783--1796},
  year={2018},
  publisher={IEEE}
}

@inproceedings{zhang2020putting,
  title={Putting visual object recognition in context},
  author={Zhang, Mengmi and Tseng, Claire and Kreiman, Gabriel},
  booktitle={Proceedings of the IEEE/CVF conference on computer vision and pattern recognition},
  pages={12985--12994},
  year={2020}
}

@inproceedings{bomatter2021pigs,
  title={When pigs fly: Contextual reasoning in synthetic and natural scenes},
  author={Bomatter, Philipp and Zhang, Mengmi and Karev, Dimitar and Madan, Spandan and Tseng, Claire and Kreiman, Gabriel},
  booktitle={Proceedings of the IEEE/CVF International Conference on Computer Vision},
  pages={255--264},
  year={2021}
}

@article{liu2022reason,
  title={Reason from context with self-supervised learning},
  author={Liu, Xiao and Sikarwar, Ankur and Kreiman, Gabriel and Shi, Zenglin and Zhang, Mengmi},
  journal={arXiv preprint arXiv:2211.12817},
  year={2022}
}

@inproceedings{tee2023integrating,
  title={Integrating curricula with replays: its effects on continual learning},
  author={Tee, Ren Jie and Zhang, Mengmi},
  booktitle={Proceedings of the AAAI Symposium Series},
  volume={1},
  number={1},
  pages={109--116},
  year={2023}
}

@inproceedings{zhang2017foveated,
  title={Foveated neural network: Gaze prediction on egocentric videos},
  author={Zhang, Mengmi and Ma, Keng Teck and Lim, Joo Hwee and Zhao, Qi},
  booktitle={2017 IEEE International Conference on Image Processing (ICIP)},
  pages={3720--3724},
  year={2017},
  organization={IEEE}
}

@article{jia2025seeing,
  title={Seeing sound, hearing sight: Uncovering modality bias and conflict of ai models in sound localization},
  author={Jia, Yanhao and Xie, Ji and Jivaganesh, S and Li, Hao and Wu, Xu and Zhang, Mengmi},
  journal={arXiv preprint arXiv:2505.11217},
  year={2025}
}

@article{talbot2023tuned,
  title={Tuned compositional feature replays for efficient stream learning},
  author={Talbot, Morgan B and Zawar, Rushikesh and Badkundri, Rohil and Zhang, Mengmi and Kreiman, Gabriel},
  journal={IEEE Transactions on Neural Networks and Learning Systems},
  volume={36},
  number={2},
  pages={3300--3314},
  year={2023},
  publisher={IEEE}
}

@article{zhang2022look,
  title={Look twice: A generalist computational model predicts return fixations across tasks and species},
  author={Zhang, Mengmi and Armendariz, Marcelo and Xiao, Will and Rose, Olivia and Bendtz, Katarina and Livingstone, Margaret and Ponce, Carlos and Kreiman, Gabriel},
  journal={PLoS computational biology},
  volume={18},
  number={11},
  pages={e1010654},
  year={2022},
  publisher={Public Library of Science San Francisco, CA USA}
}

@inproceedings{wu2023label,
  title={Label-efficient online continual object detection in streaming video},
  author={Wu, Jay Zhangjie and Zhang, David Junhao and Hsu, Wynne and Zhang, Mengmi and Shou, Mike Zheng},
  booktitle={Proceedings of the IEEE/CVF International Conference on Computer Vision},
  pages={19246--19255},
  year={2023}
}

@inproceedings{singh2023learning,
  title={Learning to learn: How to continuously teach humans and machines},
  author={Singh, Parantak and Li, You and Sikarwar, Ankur and Lei, Stan Weixian and Gao, Difei and Talbot, Morgan B and Sun, Ying and Shou, Mike Zheng and Kreiman, Gabriel and Zhang, Mengmi},
  booktitle={Proceedings of the IEEE/CVF International Conference on Computer Vision},
  pages={11708--11719},
  year={2023}
}

@article{khandelwal2023adaptive,
  title={Adaptive visual scene understanding: incremental scene graph generation},
  author={Khandelwal, Naitik and Liu, Xiao and Zhang, Mengmi},
  journal={arXiv preprint arXiv:2310.01636},
  year={2023}
}

@article{wang2023object,
  title={Object-centric learning with cyclic walks between parts and whole},
  author={Wang, Ziyu and Shou, Mike Zheng and Zhang, Mengmi},
  journal={Advances in Neural Information Processing Systems},
  volume={36},
  pages={9388--9408},
  year={2023}
}

@article{han2024flow,
  title={Flow Snapshot Neurons in Action: Deep Neural Networks Generalize to Biological Motion Perception},
  author={Han, Shuangpeng and Wang, Ziyu and Zhang, Mengmi},
  journal={Advances in Neural Information Processing Systems},
  volume={37},
  pages={53732--53763},
  year={2024}
}

@article{sikarwar2023decoding,
  title={Decoding the enigma: benchmarking humans and AIs on the many facets of working memory},
  author={Sikarwar, Ankur and Zhang, Mengmi},
  journal={Advances in Neural Information Processing Systems},
  volume={36},
  pages={74039--74076},
  year={2023}
}

@inproceedings{wang2025gazing,
  title={Gazing at Rewards: Eye Movements as a Lens into Human and AI Decision-Making in Hybrid Visual Foraging},
  author={Wang, Bo and Tan, Dingwei and Kuo, Yen-Ling and Sun, Zhaowei and Wolfe, Jeremy M and Cham, Tat-Jen and Zhang, Mengmi},
  booktitle={Proceedings of the Computer Vision and Pattern Recognition Conference},
  pages={14810--14823},
  year={2025}
}

@inproceedings{leclerc2023ffcv,
  title={FFCV: Accelerating training by removing data bottlenecks},
  author={Leclerc, Guillaume and Ilyas, Andrew and Engstrom, Logan and Park, Sung Min and Salman, Hadi and Madry, Aleksander},
  booktitle={Proceedings of the IEEE/CVF Conference on Computer Vision and Pattern Recognition},
  pages={12011--12020},
  year={2023}
}

@article{wang2024unsupervised,
  title={Pose Prior Learner: Unsupervised Categorical Prior Learning for Pose Estimation},
  author={Wang, Ziyu and Han, Shuangpeng and Zhang, Mengmi},
  journal={arXiv preprint arXiv:2410.03858},
  year={2024}
}

@article{madan2022improving,
  title={Improving generalization by mimicking the human visual diet},
  author={Madan, Spandan and Li, You and Zhang, Mengmi and Pfister, Hanspeter and Kreiman, Gabriel},
  journal={arXiv preprint arXiv:2206.07802},
  year={2022}
}

@article{zhang2025peering,
  title={Peering into the Unknown: Active View Selection with Neural Uncertainty Maps for 3D Reconstruction},
  author={Zhang, Zhengquan and Xu, Feng and Zhang, Mengmi},
  journal={arXiv preprint arXiv:2506.14856},
  year={2025}
}

@article{hendrycks2019benchmarking,
  title={Benchmarking neural network robustness to common corruptions and perturbations},
  author={Hendrycks, Dan and Dietterich, Thomas},
  journal={arXiv preprint arXiv:1903.12261},
  year={2019}
}

@inproceedings{reizenstein21co3d,
	Author = {Reizenstein, Jeremy and Shapovalov, Roman and Henzler, Philipp and Sbordone, Luca and Labatut, Patrick and Novotny, David},
	Booktitle = {International Conference on Computer Vision},
	Title = {Common Objects in 3D: Large-Scale Learning and Evaluation of Real-life 3D Category Reconstruction},
	Year = {2021},
}

@article{sullivan2021saycam,
  title={SAYCam: A large, longitudinal audiovisual dataset recorded from the infant’s perspective},
  author={Sullivan, Jessica and Mei, Michelle and Perfors, Andrew and Wojcik, Erica and Frank, Michael C},
  journal={Open mind},
  volume={5},
  pages={20--29},
  year={2021},
  publisher={MIT Press One Rogers Street, Cambridge, MA 02142-1209, USA journals-info~…}
}

@article{linsley20243d,
  title={The 3d-pc: a benchmark for visual perspective taking in humans and machines},
  author={Linsley, Drew and Zhou, Peisen and Ashok, Alekh Karkada and Nagaraj, Akash and Gaonkar, Gaurav and Lewis, Francis E and Pizlo, Zygmunt and Serre, Thomas},
  journal={arXiv preprint arXiv:2406.04138},
  year={2024}
}

@inproceedings{geirhos2018imagenet,
  title={ImageNet-trained CNNs are biased towards texture; increasing shape bias improves accuracy and robustness},
  author={Geirhos, Robert and Rubisch, Patricia and Michaelis, Claudio and Bethge, Matthias and Wichmann, Felix A and Brendel, Wieland},
  booktitle={International conference on learning representations},
  year={2018}
}

@article{ridnik2021in21k,
  title={Imagenet-21k pretraining for the masses},
  author={Ridnik, Tal and Ben-Baruch, Emanuel and Noy, Asaf and Zelnik-Manor, Lihi},
  journal={arXiv preprint arXiv:2104.10972},
  year={2021}
}

@article{rakic1986concurrent,
  title={Concurrent overproduction of synapses in diverse regions of the primate cerebral cortex},
  author={Rakic, Pasko and Bourgeois, Jean-Pierre and Eckenhoff, Maryellen F and Zecevic, Nada and Goldman-Rakic, Patricia S},
  journal={Science},
  volume={232},
  number={4747},
  pages={232--235},
  year={1986},
  publisher={American Association for the Advancement of Science}
}

@incollection{aslin2001visual,
  author    = {R. N. Aslin},
  title     = {Visual Development: Infant},
  booktitle = {International Encyclopedia of the Social \& Behavioral Sciences},
  editor    = {Neil J. Smelser and Paul B. Baltes},
  publisher = {Pergamon},
  year      = {2001},
  pages     = {16250--16255},
  isbn      = {9780080430768},
  doi       = {10.1016/B0-08-043076-7/03613-5},
  url       = {https://doi.org/10.1016/B0-08-043076-7/03613-5}
}

@article{mayer1982visual,
  title={Visual acuity development in infants and young children, as assessed by operant preferential looking},
  author={Mayer, D Luisa and Dobson, Velma},
  journal={Vision research},
  volume={22},
  number={9},
  pages={1141--1151},
  year={1982},
  publisher={Elsevier}
}

@article{norcia1985spatial,
  title={Spatial frequency sweep VEP: visual acuity during the first year of life},
  author={Norcia, Anthony M and Tyler, Christopher W},
  journal={Vision research},
  volume={25},
  number={10},
  pages={1399--1408},
  year={1985},
  publisher={Elsevier}
}

@article{teller1997first,
  title={First glances: the vision of infants. the Friedenwald lecture.},
  author={Teller, Davida Y},
  journal={Investigative ophthalmology \& visual science},
  volume={38},
  number={11},
  pages={2183--2203},
  year={1997},
  publisher={The Association for Research in Vision and Ophthalmology}
}

@article{skelton2022infant,
  title={Infant color perception: Insight into perceptual development},
  author={Skelton, Alice E and Maule, John and Franklin, Anna},
  journal={Child development perspectives},
  volume={16},
  number={2},
  pages={90--95},
  year={2022},
  publisher={Wiley Online Library}
}

@article{werner1979human,
  title={Human infant color vision and color perception},
  author={Werner, John S and Wooten, BR},
  journal={Infant Behavior and Development},
  volume={2},
  pages={241--273},
  year={1979},
  publisher={Elsevier}
}

@article{maule2023development,
  title={The development of color perception and cognition},
  author={Maule, John and Skelton, Alice E and Franklin, Anna},
  journal={Annual Review of Psychology},
  volume={74},
  number={1},
  pages={87--111},
  year={2023},
  publisher={Annual Reviews}
}

@article{dobkins2001colorspeed,
  title={Development of psychophysically-derived detection contours in L-and M-cone contrast space},
  author={Dobkins, Karen R and Anderson, Christina M and Kelly, John},
  journal={Vision Research},
  volume={41},
  number={14},
  pages={1791--1807},
  year={2001},
  publisher={Elsevier}
}

@article{crognale2002development,
  title={Development, maturation, and aging of chromatic visual pathways: VEP results},
  author={Crognale, Michael A},
  journal={Journal of Vision},
  volume={2},
  number={6},
  pages={2--2},
  year={2002},
  publisher={The Association for Research in Vision and Ophthalmology}
}

@article{gibson1960viscliff,
  title={The" visual cliff"},
  author={Gibson, Eleanor J and Walk, Richard D},
  journal={Scientific American},
  volume={202},
  number={4},
  pages={64--71},
  year={1960},
  publisher={JSTOR}
}

@article{vogelsang2018potential,
  title={Potential downside of high initial visual acuity},
  author={Vogelsang, Lukas and Gilad-Gutnick, Sharon and Ehrenberg, Evan and Yonas, Albert and Diamond, Sidney and Held, Richard and Sinha, Pawan},
  journal={Proceedings of the National Academy of Sciences},
  volume={115},
  number={44},
  pages={11333--11338},
  year={2018},
  publisher={National Academy of Sciences}
}

@article{vogelsang2024impact,
  title={Impact of early visual experience on later usage of color cues},
  author={Vogelsang, Marin and Vogelsang, Lukas and Gupta, Priti and Gandhi, Tapan K and Shah, Pragya and Swami, Piyush and Gilad-Gutnick, Sharon and Ben-Ami, Shlomit and Diamond, Sidney and Ganesh, Suma and others},
  journal={Science},
  volume={384},
  number={6698},
  pages={907--912},
  year={2024},
  publisher={American Association for the Advancement of Science}
}

@article{vogelsang2025potential,
  title={Potential role of developmental experience in the emergence of the parvo-magno distinction},
  author={Vogelsang, Marin and Vogelsang, Lukas and Pipa, Gordon and Diamond, Sidney and Sinha, Pawan},
  journal={Communications Biology},
  volume={8},
  number={1},
  pages={987},
  year={2025},
  publisher={Nature Publishing Group UK London}
}

@article{zaadnoordijk2022lessonsnature,
  title={Lessons from infant learning for unsupervised machine learning},
  author={Zaadnoordijk, Lorijn and Besold, Tarek R and Cusack, Rhodri},
  journal={Nature Machine Intelligence},
  volume={4},
  number={6},
  pages={510--520},
  year={2022},
  publisher={Nature Publishing Group UK London}
}

@article{wood2018development,
  title={The development of invariant object recognition requires visual experience with temporally smooth objects},
  author={Wood, Justin N and Wood, Samantha MW},
  journal={Cognitive Science},
  volume={42},
  number={4},
  pages={1391--1406},
  year={2018},
  publisher={Wiley Online Library}
}

@article{mckyton2015limits,
  title={The limits of shape recognition following late emergence from blindness},
  author={McKyton, Ayelet and Ben-Zion, Itay and Doron, Ravid and Zohary, Ehud},
  journal={Current Biology},
  volume={25},
  number={18},
  pages={2373--2378},
  year={2015},
  publisher={Elsevier}
}

@inproceedings{achille2018critical,
  title={Critical learning periods in deep networks},
  author={Achille, Alessandro and Rovere, Matteo and Soatto, Stefano},
  booktitle={International conference on learning representations},
  year={2018}
}

@article{dekker2025infants,
  title={How infants look shapes what they learn},
  author={Dekker, Tessa M and Maimon-Mor, Roni O},
  journal={Proceedings of the National Academy of Sciences},
  volume={122},
  number={20},
  pages={e2505492122},
  year={2025},
  publisher={National Academy of Sciences}
}

@article{smith2005development,
  title={The development of embodied cognition: Six lessons from babies},
  author={Smith, Linda and Gasser, Michael},
  journal={Artificial life},
  volume={11},
  number={1-2},
  pages={13--29},
  year={2005},
  publisher={MIT Press}
}

@article{smith2018developing,
  title={The developing infant creates a curriculum for statistical learning},
  author={Smith, Linda B and Jayaraman, Swapnaa and Clerkin, Elizabeth and Yu, Chen},
  journal={Trends in cognitive sciences},
  volume={22},
  number={4},
  pages={325--336},
  year={2018},
  publisher={Elsevier}
}

@article{smith2017developmental,
  title={A developmental approach to machine learning?},
  author={Smith, Linda B and Slone, Lauren K},
  journal={Frontiers in psychology},
  volume={8},
  pages={296143},
  year={2017},
  publisher={Frontiers}
}

@article{clerkin2017real,
  title={Real-world visual statistics and infants' first-learned object names},
  author={Clerkin, Elizabeth M and Hart, Elizabeth and Rehg, James M and Yu, Chen and Smith, Linda B},
  journal={Philosophical Transactions of the Royal Society B: Biological Sciences},
  volume={372},
  number={1711},
  pages={20160055},
  year={2017},
  publisher={The Royal Society}
}

@article{bambach2018toddler,
  title={Toddler-inspired visual object learning},
  author={Bambach, Sven and Crandall, David and Smith, Linda and Yu, Chen},
  journal={Advances in neural information processing systems},
  volume={31},
  year={2018}
}

@article{blakemore1970development,
  title={Development of the brain depends on the visual environment},
  author={Blakemore, Colin and Cooper, Grahame F},
  journal={Nature},
  volume={228},
  number={5270},
  pages={477--478},
  year={1970},
  publisher={Nature Publishing Group UK London}
}

@article{ayzenberg2025fast,
  title={Fast and robust visual object recognition in young children},
  author={Ayzenberg, Vladislav and Sener, Sukran Bahar and Novick, Kylee and Lourenco, Stella F},
  journal={Science Advances},
  volume={11},
  number={27},
  pages={eads6821},
  year={2025},
  publisher={American Association for the Advancement of Science}
}

@article{vogelsang2024butterfly,
  title={Butterfly effects in perceptual development: A review of the ‘adaptive initial degradation’hypothesis},
  author={Vogelsang, Lukas and Vogelsang, Marin and Pipa, Gordon and Diamond, Sidney and Sinha, Pawan},
  journal={Developmental Review},
  volume={71},
  pages={101117},
  year={2024},
  publisher={Elsevier}
}

@article{geldart2002effect,
  title={The effect of early visual deprivation on the development of face processing},
  author={Geldart, Sybil and Mondloch, Catherine J and Maurer, Daphne and De Schonen, Scania and Brent, Henry P},
  journal={Developmental Science},
  volume={5},
  number={4},
  pages={490--501},
  year={2002},
  publisher={Wiley Online Library}
}

@article{maurer2007sleeper,
  title={Sleeper effects},
  author={Maurer, Daphne and Mondloch, Catherine J and Lewis, Terri L},
  journal={Developmental Science},
  volume={10},
  number={1},
  pages={40--47},
  year={2007},
  publisher={Wiley Online Library}
}

@article{putzar2010early,
  title={Early visual deprivation affects the development of face recognition and of audio-visual speech perception},
  author={Putzar, Lisa and H{\"o}tting, Kirsten and R{\"o}der, Brigitte},
  journal={Restorative neurology and neuroscience},
  volume={28},
  number={2},
  pages={251--257},
  year={2010},
  publisher={SAGE Publications Sage UK: London, England}
}

@article{dominguez2003binocular,
  title={Developmental constraints aid the acquisition of binocular disparity sensitivities},
  author={Dominguez, Melissa and Jacobs, Robert A},
  journal={Neural Computation},
  volume={15},
  number={1},
  pages={161--182},
  year={2003},
  publisher={MIT Press}
}

@article{schwarzer2014motor,
  title={How motor and visual experiences shape infants' visual processing of objects and faces},
  author={Schwarzer, Gudrun},
  journal={Child Development Perspectives},
  volume={8},
  number={4},
  pages={213--217},
  year={2014},
  publisher={Wiley Online Library}
}

@article{johnson2010infantsmotion,
  title={How infants learn about the visual world},
  author={Johnson, Scott P},
  journal={Cognitive science},
  volume={34},
  number={7},
  pages={1158--1184},
  year={2010},
  publisher={Wiley Online Library}
}

@article{myowa2011goal,
  title={Visual experience influences 12-month-old infants' perception of goal-directed actions of others.},
  author={Myowa-Yamakoshi, Masako and Kawakita, Yuka and Okanda, Mako and Takeshita, Hideko},
  journal={Developmental psychology},
  volume={47},
  number={4},
  pages={1042},
  year={2011},
  publisher={American Psychological Association}
}

@article{moulson2011neural,
  title={Neural correlates of visual recognition in 3-month-old infants: The role of experience},
  author={Moulson, Margaret C and Shannon, Robert W and Nelson, Charles A},
  journal={Developmental psychobiology},
  volume={53},
  number={4},
  pages={416--424},
  year={2011},
  publisher={Wiley Online Library}
}

@article{arcaro2017monkeyface,
  title={Seeing faces is necessary for face-domain formation},
  author={Arcaro, Michael J and Schade, Peter F and Vincent, Justin L and Ponce, Carlos R and Livingstone, Margaret S},
  journal={Nature neuroscience},
  volume={20},
  number={10},
  pages={1404--1412},
  year={2017},
  publisher={Nature Publishing Group US New York}
}

@article{petroff2025world,
  title={The world through infant eyes: Evidence for the early emergence of the cardinal orientation bias},
  author={Petroff, Zachary J and Jayaraman, Swapnaa and Smith, Linda B and Candy, T Rowan and Bonnen, Kathryn},
  journal={Proceedings of the National Academy of Sciences},
  volume={122},
  number={16},
  pages={e2421277122},
  year={2025},
  publisher={National Academy of Sciences}
}

@article{franchak2024smith,
  title={Developmental changes in how head orientation structures infants’ visual attention},
  author={Franchak, John M and Smith, Linda and Yu, Chen},
  journal={Developmental psychobiology},
  volume={66},
  number={7},
  pages={e22538},
  year={2024},
  publisher={Wiley Online Library}
}

@article{orhan2020self,
  title={Self-supervised learning through the eyes of a child},
  author={Orhan, Emin and Gupta, Vaibhav and Lake, Brenden M},
  journal={Advances in Neural Information Processing Systems},
  volume={33},
  pages={9960--9971},
  year={2020}
}

@article{sheybani2023curriculum,
  title={Curriculum learning with infant egocentric videos},
  author={Sheybani, Saber and Hansaria, Himanshu and Wood, Justin and Smith, Linda and Tiganj, Zoran},
  journal={Advances in Neural Information Processing Systems},
  volume={36},
  pages={54199--54212},
  year={2023}
}

@article{orhan2024learning,
  title={Learning high-level visual representations from a child’s perspective without strong inductive biases},
  author={Orhan, A Emin and Lake, Brenden M},
  journal={Nature Machine Intelligence},
  volume={6},
  number={3},
  pages={271--283},
  year={2024},
  publisher={Nature Publishing Group UK London}
}

@inproceedings{zhang2019makingrf1,
  title={Making convolutional networks shift-invariant again},
  author={Zhang, Richard},
  booktitle={International conference on machine learning},
  pages={7324--7334},
  year={2019},
  organization={PMLR}
}

@inproceedings{chaman2021trulyrf2,
  title={Truly shift-invariant convolutional neural networks},
  author={Chaman, Anadi and Dokmanic, Ivan},
  booktitle={Proceedings of the IEEE/CVF Conference on Computer Vision and Pattern Recognition},
  pages={3773--3783},
  year={2021}
}

@article{madan2021smallrf3fool,
  title={Small in-distribution changes in 3d perspective and lighting fool both cnns and transformers},
  author={Madan, Spandan and Sasaki, Tomotake and Li, Tzu-Mao and Boix, Xavier and Pfister, Hanspeter},
  journal={arXiv preprint arXiv:2106.16198},
  volume={3},
  year={2021}
}

@inproceedings{beery2018recognitionrf4,
  title={Recognition in terra incognita},
  author={Beery, Sara and Van Horn, Grant and Perona, Pietro},
  booktitle={Proceedings of the European conference on computer vision (ECCV)},
  pages={456--473},
  year={2018}
}

@article{zhang2024adversarialrf5,
  title={Adversarial relighting against face recognition},
  author={Zhang, Qian and Guo, Qing and Gao, Ruijun and Juefei-Xu, Felix and Yu, Hongkai and Feng, Wei},
  journal={IEEE Transactions on Information Forensics and Security},
  volume={19},
  pages={9145--9157},
  year={2024},
  publisher={IEEE}
}

@article{barbu2019objectnetrf6,
  title={Objectnet: A large-scale bias-controlled dataset for pushing the limits of object recognition models},
  author={Barbu, Andrei and Mayo, David and Alverio, Julian and Luo, William and Wang, Christopher and Gutfreund, Dan and Tenenbaum, Josh and Katz, Boris},
  journal={Advances in neural information processing systems},
  volume={32},
  year={2019}
}

@article{liu2018beyondrf7,
  title={Beyond pixel norm-balls: Parametric adversaries using an analytically differentiable renderer},
  author={Liu, Hsueh-Ti Derek and Tao, Michael and Li, Chun-Liang and Nowrouzezahrai, Derek and Jacobson, Alec},
  journal={arXiv preprint arXiv:1808.02651},
  year={2018}
}

@inproceedings{zeng2019adversarialrf8,
  title={Adversarial attacks beyond the image space},
  author={Zeng, Xiaohui and Liu, Chenxi and Wang, Yu-Siang and Qiu, Weichao and Xie, Lingxi and Tai, Yu-Wing and Tang, Chi-Keung and Yuille, Alan L},
  booktitle={Proceedings of the IEEE/CVF Conference on Computer Vision and Pattern Recognition},
  pages={4302--4311},
  year={2019}
}

@article{sakai2022threerf9,
  title={Three approaches to facilitate invariant neurons and generalization to out-of-distribution orientations and illuminations},
  author={Sakai, Akira and Sunagawa, Taro and Madan, Spandan and Suzuki, Kanata and Katoh, Takashi and Kobashi, Hiromichi and Pfister, Hanspeter and Sinha, Pawan and Boix, Xavier and Sasaki, Tomotake},
  journal={Neural Networks},
  volume={155},
  pages={119--143},
  year={2022},
  publisher={Elsevier}
}

@inproceedings{joshi2019semanticrf10fool,
  title={Semantic adversarial attacks: Parametric transformations that fool deep classifiers},
  author={Joshi, Ameya and Mukherjee, Amitangshu and Sarkar, Soumik and Hegde, Chinmay},
  booktitle={Proceedings of the IEEE/CVF international conference on computer vision},
  pages={4773--4783},
  year={2019}
}

@inproceedings{yun2019cutmixrf12,
  title={Cutmix: Regularization strategy to train strong classifiers with localizable features},
  author={Yun, Sangdoo and Han, Dongyoon and Oh, Seong Joon and Chun, Sanghyuk and Choe, Junsuk and Yoo, Youngjoon},
  booktitle={Proceedings of the IEEE/CVF international conference on computer vision},
  pages={6023--6032},
  year={2019}
}

@article{hendrycks2019augmixrf13,
  title={Augmix: A simple data processing method to improve robustness and uncertainty},
  author={Hendrycks, Dan and Mu, Norman and Cubuk, Ekin D and Zoph, Barret and Gilmer, Justin and Lakshminarayanan, Balaji},
  journal={International Conference on Learning Representations},
  year={2019}
}

@article{zhang2017mixuprf14,
  title={mixup: Beyond empirical risk minimization},
  author={Zhang, Hongyi and Cisse, Moustapha and Dauphin, Yann N and Lopez-Paz, David},
  journal={International Conference on Learning Representations},
  year={2017}
}

@inproceedings{ilse2020divarf15,
  title={Diva: Domain invariant variational autoencoders},
  author={Ilse, Maximilian and Tomczak, Jakub M and Louizos, Christos and Welling, Max},
  booktitle={Medical Imaging with Deep Learning},
  pages={322--348},
  year={2020},
  organization={PMLR}
}

@inproceedings{wang2020rf16,
  title={Cross-domain face presentation attack detection via multi-domain disentangled representation learning},
  author={Wang, Guoqing and Han, Hu and Shan, Shiguang and Chen, Xilin},
  booktitle={Proceedings of the IEEE/CVF conference on computer vision and pattern recognition},
  pages={6678--6687},
  year={2020}
}

@inproceedings{muandet2013domainrf17,
  title={Domain generalization via invariant feature representation},
  author={Muandet, Krikamol and Balduzzi, David and Sch{\"o}lkopf, Bernhard},
  booktitle={International conference on machine learning},
  pages={10--18},
  year={2013},
  organization={PMLR}
}

@inproceedings{li2018domainrf18,
  title={Domain generalization with adversarial feature learning},
  author={Li, Haoliang and Pan, Sinno Jialin and Wang, Shiqi and Kot, Alex C},
  booktitle={Proceedings of the IEEE conference on computer vision and pattern recognition},
  pages={5400--5409},
  year={2018}
}

@inproceedings{li2018deeprf19,
  title={Deep domain generalization via conditional invariant adversarial networks},
  author={Li, Ya and Tian, Xinmei and Gong, Mingming and Liu, Yajing and Liu, Tongliang and Zhang, Kun and Tao, Dacheng},
  booktitle={Proceedings of the European conference on computer vision (ECCV)},
  pages={624--639},
  year={2018}
}

@inproceedings{motiian2017unifiedrf20,
  title={Unified deep supervised domain adaptation and generalization},
  author={Motiian, Saeid and Piccirilli, Marco and Adjeroh, Donald A and Doretto, Gianfranco},
  booktitle={Proceedings of the IEEE international conference on computer vision},
  pages={5715--5725},
  year={2017}
}

@inproceedings{shao2019multirf21,
  title={Multi-adversarial discriminative deep domain generalization for face presentation attack detection},
  author={Shao, Rui and Lan, Xiangyuan and Li, Jiawei and Yuen, Pong C},
  booktitle={Proceedings of the IEEE/CVF conference on computer vision and pattern recognition},
  pages={10023--10031},
  year={2019}
}

@inproceedings{wang2021respectingrf22,
  title={Respecting domain relations: Hypothesis invariance for domain generalization},
  author={Wang, Ziqi and Loog, Marco and Van Gemert, Jan},
  booktitle={2020 25th International Conference on Pattern Recognition (ICPR)},
  pages={9756--9763},
  year={2021},
  organization={IEEE}
}

@article{shahtalebi2021sandrf23,
  title={Sand-mask: An enhanced gradient masking strategy for the discovery of invariances in domain generalization},
  author={Shahtalebi, Soroosh and Gagnon-Audet, Jean-Christophe and Laleh, Touraj and Faramarzi, Mojtaba and Ahuja, Kartik and Rish, Irina},
  journal={arXiv preprint arXiv:2106.02266},
  year={2021}
}

@inproceedings{sun2016deeprf24,
  title={Deep coral: Correlation alignment for deep domain adaptation},
  author={Sun, Baochen and Saenko, Kate},
  booktitle={European conference on computer vision},
  pages={443--450},
  year={2016},
  organization={Springer}
}

@article{arjovsky2019invariantrf25,
  title={Invariant risk minimization},
  author={Arjovsky, Martin and Bottou, L{\'e}on and Gulrajani, Ishaan and Lopez-Paz, David},
  journal={arXiv preprint arXiv:1907.02893},
  year={2019}
}

@inproceedings{kim2021selfregrf26,
  title={Selfreg: Self-supervised contrastive regularization for domain generalization},
  author={Kim, Daehee and Yoo, Youngjun and Park, Seunghyun and Kim, Jinkyu and Lee, Jaekoo},
  booktitle={Proceedings of the IEEE/CVF international conference on computer vision},
  pages={9619--9628},
  year={2021}
}

@article{vedantam2021empiricalrf27,
  title={An empirical investigation of domain generalization with empirical risk minimizers},
  author={Vedantam, Ramakrishna and Lopez-Paz, David and Schwab, David J},
  journal={Advances in neural information processing systems},
  volume={34},
  pages={28131--28143},
  year={2021}
}

@inproceedings{krueger2021outrf28,
  title={Out-of-distribution generalization via risk extrapolation (rex)},
  author={Krueger, David and Caballero, Ethan and Jacobsen, Joern-Henrik and Zhang, Amy and Binas, Jonathan and Zhang, Dinghuai and Le Priol, Remi and Courville, Aaron},
  booktitle={International conference on machine learning},
  pages={5815--5826},
  year={2021},
  organization={PMLR}
}

@article{blanchard2021domainrf29,
  title={Domain generalization by marginal transfer learning},
  author={Blanchard, Gilles and Deshmukh, Aniket Anand and Dogan, Urun and Lee, Gyemin and Scott, Clayton},
  journal={Journal of machine learning research},
  volume={22},
  number={2},
  pages={1--55},
  year={2021}
}

@article{ren2019likelihoodrf30,
  title={Likelihood ratios for out-of-distribution detection},
  author={Ren, Jie and Liu, Peter J and Fertig, Emily and Snoek, Jasper and Poplin, Ryan and Depristo, Mark and Dillon, Joshua and Lakshminarayanan, Balaji},
  journal={Advances in neural information processing systems},
  volume={32},
  year={2019}
}

@article{sastry2019detectingrf31,
  title={Detecting out-of-distribution examples with in-distribution examples and gram matrices},
  author={Sastry, Chandramouli Shama and Oore, Sageev},
  journal={arXiv preprint arXiv:1912.12510},
  year={2019}
}

@article{hodge2004surveyrf32,
  title={A survey of outlier detection methodologies},
  author={Hodge, Victoria and Austin, Jim},
  journal={Artificial intelligence review},
  volume={22},
  number={2},
  pages={85--126},
  year={2004},
  publisher={Springer}
}

@inproceedings{aggarwal2001outlierrf33,
  title={Outlier detection for high dimensional data},
  author={Aggarwal, Charu C and Yu, Philip S},
  booktitle={Proceedings of the 2001 ACM SIGMOD international conference on Management of data},
  pages={37--46},
  year={2001}
}

@inproceedings{hendrycks2021many,
  title={The many faces of robustness: A critical analysis of out-of-distribution generalization},
  author={Hendrycks, Dan and Basart, Steven and Mu, Norman and Kadavath, Saurav and Wang, Frank and Dorundo, Evan and Desai, Rahul and Zhu, Tyler and Parajuli, Samyak and Guo, Mike and others},
  booktitle={Proceedings of the IEEE/CVF international conference on computer vision},
  pages={8340--8349},
  year={2021}
}

@article{mintun2021interaction,
  title={On interaction between augmentations and corruptions in natural corruption robustness},
  author={Mintun, Eric and Kirillov, Alexander and Xie, Saining},
  journal={Advances in Neural Information Processing Systems},
  volume={34},
  pages={3571--3583},
  year={2021}
}

@article{wang2023survey,
  title={A survey on the robustness of computer vision models against common corruptions},
  author={Wang, Shunxin and Veldhuis, Raymond and Brune, Christoph and Strisciuglio, Nicola},
  journal={arXiv preprint arXiv:2305.06024},
  year={2023}
}

@inproceedings{schneider2021simclrtt,
  title={Contrastive learning through time},
  author={Schneider, Felix and Xu, Xia and Ernst, Markus R and Yu, Zhengyang and Triesch, Jochen},
  booktitle={SVRHM 2021 Workshop@ NeurIPS},
  year={2021}
}

@inproceedings{he2022masked,
  title={Masked autoencoders are scalable vision learners},
  author={He, Kaiming and Chen, Xinlei and Xie, Saining and Li, Yanghao and Doll{\'a}r, Piotr and Girshick, Ross},
  booktitle={Proceedings of the IEEE/CVF conference on computer vision and pattern recognition},
  pages={16000--16009},
  year={2022}
}

@article{halvagal2023combination,
  title={The combination of Hebbian and predictive plasticity learns invariant object representations in deep sensory networks},
  author={Halvagal, Manu Srinath and Zenke, Friedemann},
  journal={Nature neuroscience},
  volume={26},
  number={11},
  pages={1906--1915},
  year={2023},
  publisher={Nature Publishing Group US New York}
}

@inproceedings{chen2020simple,
  title={A simple framework for contrastive learning of visual representations},
  author={Chen, Ting and Kornblith, Simon and Norouzi, Mohammad and Hinton, Geoffrey},
  booktitle={International conference on machine learning},
  pages={1597--1607},
  year={2020},
  organization={PmLR}
}

@inproceedings{he2016deep,
  title={Deep residual learning for image recognition},
  author={He, Kaiming and Zhang, Xiangyu and Ren, Shaoqing and Sun, Jian},
  booktitle={Proceedings of the IEEE conference on computer vision and pattern recognition},
  pages={770--778},
  year={2016}
}

@article{dosovitskiy2020image,
  title={An image is worth 16x16 words: Transformers for image recognition at scale},
  author={Dosovitskiy, Alexey},
  journal={International Conference on Learning Representations},
  year={2020}
}

@inproceedings{caron2021emerging,
  title={Emerging properties in self-supervised vision transformers},
  author={Caron, Mathilde and Touvron, Hugo and Misra, Ishan and J{\'e}gou, Herv{\'e} and Mairal, Julien and Bojanowski, Piotr and Joulin, Armand},
  booktitle={Proceedings of the IEEE/CVF international conference on computer vision},
  pages={9650--9660},
  year={2021}
}

@article{oquab2023dinov2,
  title={Dinov2: Learning robust visual features without supervision},
  author={Oquab, Maxime and Darcet, Timoth{\'e}e and Moutakanni, Th{\'e}o and Vo, Huy and Szafraniec, Marc and Khalidov, Vasil and Fernandez, Pierre and Haziza, Daniel and Massa, Francisco and El-Nouby, Alaaeldin and others},
  journal={Transactions on Machine Learning Research},
  year={2023}
}

@article{hinton1993autoencoders,
  title={Autoencoders, minimum description length and Helmholtz free energy},
  author={Hinton, Geoffrey E and Zemel, Richard},
  journal={Advances in neural information processing systems},
  volume={6},
  year={1993}
}

@article{tong2022videomae,
  title={Videomae: Masked autoencoders are data-efficient learners for self-supervised video pre-training},
  author={Tong, Zhan and Song, Yibing and Wang, Jue and Wang, Limin},
  journal={Advances in neural information processing systems},
  volume={35},
  pages={10078--10093},
  year={2022}
}

@inproceedings{wang2023videomae2,
  title={Videomae v2: Scaling video masked autoencoders with dual masking},
  author={Wang, Limin and Huang, Bingkun and Zhao, Zhiyu and Tong, Zhan and He, Yinan and Wang, Yi and Wang, Yali and Qiao, Yu},
  booktitle={Proceedings of the IEEE/CVF conference on computer vision and pattern recognition},
  pages={14549--14560},
  year={2023}
}

@inproceedings{wu2018InstDisc,
  title={Unsupervised feature learning via non-parametric instance discrimination},
  author={Wu, Zhirong and Xiong, Yuanjun and Yu, Stella X and Lin, Dahua},
  booktitle={Proceedings of the IEEE conference on computer vision and pattern recognition},
  pages={3733--3742},
  year={2018}
}

@inproceedings{ye2019invaspread,
  title={Unsupervised embedding learning via invariant and spreading instance feature},
  author={Ye, Mang and Zhang, Xu and Yuen, Pong C and Chang, Shih-Fu},
  booktitle={Proceedings of the IEEE/CVF conference on computer vision and pattern recognition},
  pages={6210--6219},
  year={2019}
}

@article{chen2020simclrv2,
  title={Big self-supervised models are strong semi-supervised learners},
  author={Chen, Ting and Kornblith, Simon and Swersky, Kevin and Norouzi, Mohammad and Hinton, Geoffrey E},
  journal={Advances in neural information processing systems},
  volume={33},
  pages={22243--22255},
  year={2020}
}

@article{oord2018cpcv1,
  title={Representation learning with contrastive predictive coding},
  author={Oord, Aaron van den and Li, Yazhe and Vinyals, Oriol},
  journal={arXiv preprint arXiv:1807.03748},
  year={2018}
}

@inproceedings{henaff2020cpcv2,
  title={Data-efficient image recognition with contrastive predictive coding},
  author={Henaff, Olivier},
  booktitle={International conference on machine learning},
  pages={4182--4192},
  year={2020},
  organization={PMLR}
}

@inproceedings{he2020momentummoco,
  title={Momentum contrast for unsupervised visual representation learning},
  author={He, Kaiming and Fan, Haoqi and Wu, Yuxin and Xie, Saining and Girshick, Ross},
  booktitle={Proceedings of the IEEE/CVF conference on computer vision and pattern recognition},
  pages={9729--9738},
  year={2020}
}

@article{chen2020improvedmocov2,
  title={Improved baselines with momentum contrastive learning},
  author={Chen, Xinlei and Fan, Haoqi and Girshick, Ross and He, Kaiming},
  journal={arXiv preprint arXiv:2003.04297},
  year={2020}
}

@inproceedings{chen2021empiricalmocov3,
  title={An empirical study of training self-supervised vision transformers},
  author={Chen, Xinlei and Xie, Saining and He, Kaiming},
  booktitle={Proceedings of the IEEE/CVF international conference on computer vision},
  pages={9640--9649},
  year={2021}
}

@inproceedings{chen2021simsam,
  title={Exploring simple siamese representation learning},
  author={Chen, Xinlei and He, Kaiming},
  booktitle={Proceedings of the IEEE/CVF conference on computer vision and pattern recognition},
  pages={15750--15758},
  year={2021}
}

@article{grill2020byol,
  title={Bootstrap your own latent-a new approach to self-supervised learning},
  author={Grill, Jean-Bastien and Strub, Florian and Altch{\'e}, Florent and Tallec, Corentin and Richemond, Pierre and Buchatskaya, Elena and Doersch, Carl and Avila Pires, Bernardo and Guo, Zhaohan and Gheshlaghi Azar, Mohammad and others},
  journal={Advances in neural information processing systems},
  volume={33},
  pages={21271--21284},
  year={2020}
}

@inproceedings{zbontar2021barlow,
  title={Barlow twins: Self-supervised learning via redundancy reduction},
  author={Zbontar, Jure and Jing, Li and Misra, Ishan and LeCun, Yann and Deny, St{\'e}phane},
  booktitle={International conference on machine learning},
  pages={12310--12320},
  year={2021},
  organization={PMLR}
}

@article{caron2020swav,
  title={Unsupervised learning of visual features by contrasting cluster assignments},
  author={Caron, Mathilde and Misra, Ishan and Mairal, Julien and Goyal, Priya and Bojanowski, Piotr and Joulin, Armand},
  journal={Advances in neural information processing systems},
  volume={33},
  pages={9912--9924},
  year={2020}
}

@article{simeoni2025dinov3,
  title={Dinov3},
  author={Sim{\'e}oni, Oriane and Vo, Huy V and Seitzer, Maximilian and Baldassarre, Federico and Oquab, Maxime and Jose, Cijo and Khalidov, Vasil and Szafraniec, Marc and Yi, Seungeun and Ramamonjisoa, Micha{\"e}l and others},
  journal={arXiv preprint arXiv:2508.10104},
  year={2025}
}

@article{bardes2021vicregbarlow,
  title={Vicreg: Variance-invariance-covariance regularization for self-supervised learning},
  author={Bardes, Adrien and Ponce, Jean and LeCun, Yann},
  journal={arXiv preprint arXiv:2105.04906},
  year={2021}
}

@inproceedings{koohpayegani2021meanbyol,
  title={Mean shift for self-supervised learning},
  author={Koohpayegani, Soroush Abbasi and Tejankar, Ajinkya and Pirsiavash, Hamed},
  booktitle={Proceedings of the IEEE/CVF International Conference on Computer Vision},
  pages={10326--10335},
  year={2021}
}

@inproceedings{ermolov2021whiteningwmse,
  title={Whitening for self-supervised representation learning},
  author={Ermolov, Aleksandr and Siarohin, Aliaksandr and Sangineto, Enver and Sebe, Nicu},
  booktitle={International conference on machine learning},
  pages={3015--3024},
  year={2021},
  organization={PMLR}
}

@inproceedings{kim2022selfdino,
  title={Self-taught metric learning without labels},
  author={Kim, Sungyeon and Kim, Dongwon and Cho, Minsu and Kwak, Suha},
  booktitle={Proceedings of the IEEE/CVF conference on computer vision and pattern recognition},
  pages={7431--7441},
  year={2022}
}

@inproceedings{qian2021spatiotemporalvideocl1,
  title={Spatiotemporal contrastive video representation learning},
  author={Qian, Rui and Meng, Tianjian and Gong, Boqing and Yang, Ming-Hsuan and Wang, Huisheng and Belongie, Serge and Cui, Yin},
  booktitle={Proceedings of the IEEE/CVF conference on computer vision and pattern recognition},
  pages={6964--6974},
  year={2021}
}

@article{han2020videocl2,
  title={Self-supervised co-training for video representation learning},
  author={Han, Tengda and Xie, Weidi and Zisserman, Andrew},
  journal={Advances in neural information processing systems},
  volume={33},
  pages={5679--5690},
  year={2020}
}

@inproceedings{feichtenhofer2021videocl3,
  title={A large-scale study on unsupervised spatiotemporal representation learning},
  author={Feichtenhofer, Christoph and Fan, Haoqi and Xiong, Bo and Girshick, Ross and He, Kaiming},
  booktitle={Proceedings of the IEEE/CVF conference on computer vision and pattern recognition},
  pages={3299--3309},
  year={2021}
}

@article{wang2024poodlevideocl4,
  title={PooDLe: Pooled and dense self-supervised learning from naturalistic videos},
  author={Wang, Alex N and Hoang, Christopher and Xiong, Yuwen and LeCun, Yann and Ren, Mengye},
  journal={arXiv preprint arXiv:2408.11208},
  year={2024}
}

@inproceedings{kuang2021videocl9,
  title={Video contrastive learning with global context},
  author={Kuang, Haofei and Zhu, Yi and Zhang, Zhi and Li, Xinyu and Tighe, Joseph and Schwertfeger, S{\"o}ren and Stachniss, Cyrill and Li, Mu},
  booktitle={Proceedings of the IEEE/CVF International Conference on Computer Vision},
  pages={3195--3204},
  year={2021}
}

@inproceedings{aubret2024interactionvideocl5,
  title={Self-supervised visual learning from interactions with objects},
  author={Aubret, Arthur and Teuli{\`e}re, C{\'e}line and Triesch, Jochen},
  booktitle={European Conference on Computer Vision},
  pages={54--71},
  year={2024},
  organization={Springer}
}

@inproceedings{wu2021contrastivevideocl7,
  title={Contrastive learning of image representations with cross-video cycle-consistency},
  author={Wu, Haiping and Wang, Xiaolong},
  booktitle={Proceedings of the IEEE/CVF International Conference on Computer Vision},
  pages={10149--10159},
  year={2021}
}

@inproceedings{tschannen2020selfvideocl8,
  title={Self-supervised learning of video-induced visual invariances},
  author={Tschannen, Michael and Djolonga, Josip and Ritter, Marvin and Mahendran, Aravindh and Houlsby, Neil and Gelly, Sylvain and Lucic, Mario},
  booktitle={Proceedings of the IEEE/CVF Conference on Computer Vision and Pattern Recognition},
  pages={13806--13815},
  year={2020}
}

@article{parthasarathy2023inavideocl6,
  title={Self-supervised video pretraining yields robust and more human-aligned visual representations},
  author={Parthasarathy, Nikhil and Eslami, SM and Carreira, Joao and Henaff, Olivier},
  journal={Advances in Neural Information Processing Systems},
  volume={36},
  pages={65743--65765},
  year={2023}
}

@article{nguyen2021robot1,
  title={Robust safety-critical control for dynamic robotics},
  author={Nguyen, Quan and Sreenath, Koushil},
  journal={IEEE Transactions on Automatic Control},
  volume={67},
  number={3},
  pages={1073--1088},
  year={2021},
  publisher={IEEE}
}

@article{zhang2024robot2,
  title={Toward robust robot 3-d perception in urban environments: The ut campus object dataset},
  author={Zhang, Arthur and Eranki, Chaitanya and Zhang, Christina and Park, Ji-Hwan and Hong, Raymond and Kalyani, Pranav and Kalyanaraman, Lochana and Gamare, Arsh and Bagad, Arnav and Esteva, Maria and others},
  journal={IEEE Transactions on Robotics},
  volume={40},
  pages={3322--3340},
  year={2024},
  publisher={IEEE}
}

@article{wang2024driving1,
  title={Explainable deep adversarial reinforcement learning approach for robust autonomous driving},
  author={Wang, Chuyao and Aouf, Nabil},
  journal={IEEE Transactions on Intelligent Vehicles},
  year={2024},
  publisher={IEEE}
}

@article{xie2025driving2,
  title={Benchmarking and improving bird's eye view perception robustness in autonomous driving},
  author={Xie, Shaoyuan and Kong, Lingdong and Zhang, Wenwei and Ren, Jiawei and Pan, Liang and Chen, Kai and Liu, Ziwei},
  journal={IEEE Transactions on Pattern Analysis and Machine Intelligence},
  year={2025},
  publisher={IEEE}
}

@article{zhao2024driving3,
  title={Enhancing autonomous driving safety: A robust traffic sign detection and recognition model TSD-YOLO},
  author={Zhao, Ruixin and Tang, Sai Hong and Shen, Jiazheng and Supeni, Eris Elianddy Bin and Rahim, Sharafiz Abdul},
  journal={Signal Processing},
  volume={225},
  pages={109619},
  year={2024},
  publisher={Elsevier}
}

@article{xu2024embodied1,
  title={From perfect to noisy world simulation: Customizable embodied multi-modal perturbations for slam robustness benchmarking},
  author={Xu, Xiaohao and Zhang, Tianyi and Wang, Sibo and Li, Xiang and Chen, Yongqi and Li, Ye and Raj, Bhiksha and Johnson-Roberson, Matthew and Huang, Xiaonan},
  journal={arXiv preprint arXiv:2406.16850},
  year={2024}
}

@article{yang2025embodied2,
  title={Reinforced Embodied Active Defense: Exploiting Adaptive Interaction for Robust Visual Perception in Adversarial 3D Environments},
  author={Yang, Xiao and Wu, Lingxuan and Wang, Lizhong and Ying, Chengyang and Su, Hang and Zhu, Jun},
  journal={IEEE Transactions on Pattern Analysis and Machine Intelligence},
  year={2025},
  publisher={IEEE}
}

@article{rebuffi2021data,
  title={Data augmentation can improve robustness},
  author={Rebuffi, Sylvestre-Alvise and Gowal, Sven and Calian, Dan Andrei and Stimberg, Florian and Wiles, Olivia and Mann, Timothy A},
  journal={Advances in neural information processing systems},
  volume={34},
  pages={29935--29948},
  year={2021}
}

@inproceedings{zhong2020random,
  title={Random erasing data augmentation},
  author={Zhong, Zhun and Zheng, Liang and Kang, Guoliang and Li, Shaozi and Yang, Yi},
  booktitle={Proceedings of the AAAI conference on artificial intelligence},
  volume={34},
  number={07},
  pages={13001--13008},
  year={2020}
}

@article{chen2020gridmask,
  title={Gridmask data augmentation},
  author={Chen, Pengguang and Liu, Shu and Zhao, Hengshuang and Wang, Xingquan and Jia, Jiaya},
  journal={arXiv preprint arXiv:2001.04086},
  year={2020}
}

@inproceedings{cubuk2019autoaugment,
  title={Autoaugment: Learning augmentation strategies from data},
  author={Cubuk, Ekin D and Zoph, Barret and Mane, Dandelion and Vasudevan, Vijay and Le, Quoc V},
  booktitle={Proceedings of the IEEE/CVF conference on computer vision and pattern recognition},
  pages={113--123},
  year={2019}
}

@inproceedings{singh2024synthetic,
  title={Is synthetic data all we need? benchmarking the robustness of models trained with synthetic images},
  author={Singh, Krishnakant and Navaratnam, Thanush and Holmer, Jannik and Schaub-Meyer, Simone and Roth, Stefan},
  booktitle={Proceedings of the IEEE/CVF Conference on Computer Vision and Pattern Recognition},
  pages={2505--2515},
  year={2024}
}

@article{vasiljevic2016blur,
  title={Examining the impact of blur on recognition by convolutional networks},
  author={Vasiljevic, Igor and Chakrabarti, Ayan and Shakhnarovich, Gregory},
  journal={arXiv preprint arXiv:1611.05760},
  year={2016}
}

@article{chiu2022colormachine,
  title={On human visual contrast sensitivity and machine vision robustness: A comparative study},
  author={Chiu, Ming-Chang and Wang, Yingfei and Kim, Derrick Eui Gyu and Chen, Pin-Yu and Ma, Xuezhe},
  journal={arXiv preprint arXiv:2212.08650},
  volume={1},
  number={3},
  year={2022}
}

@inproceedings{dodge2016quality,
  title={Understanding how image quality affects deep neural networks},
  author={Dodge, Samuel and Karam, Lina},
  booktitle={2016 eighth international conference on quality of multimedia experience (QoMEX)},
  pages={1--6},
  year={2016},
  organization={IEEE}
}

@article{yamamoto2025dinoplausible,
  title={Emergence of human-like attention and distinct head clusters in self-supervised vision transformers: A comparative eye-tracking study},
  author={Yamamoto, Takuto and Akahoshi, Hirosato and Kitazawa, Shigeru},
  journal={Neural Networks},
  pages={107595},
  year={2025},
  publisher={Elsevier}
}

@article{yerxa2024contrastiveplausible,
  title={Contrastive-equivariant self-supervised learning improves alignment with primate visual area it},
  author={Yerxa, Thomas and Feather, Jenelle and Simoncelli, Eero and Chung, SueYeon},
  journal={Advances in neural information processing systems},
  volume={37},
  pages={96045--96070},
  year={2024}
}

@article{zhuang2021simclrplausible1,
  title={Unsupervised neural network models of the ventral visual stream},
  author={Zhuang, Chengxu and Yan, Siming and Nayebi, Aran and Schrimpf, Martin and Frank, Michael C and DiCarlo, James J and Yamins, Daniel LK},
  journal={Proceedings of the National Academy of Sciences},
  volume={118},
  number={3},
  pages={e2014196118},
  year={2021},
  publisher={National Academy of Sciences}
}

@article{konkle2022simclrplausible2,
  title={A self-supervised domain-general learning framework for human ventral stream representation},
  author={Konkle, Talia and Alvarez, George A},
  journal={Nature communications},
  volume={13},
  number={1},
  pages={491},
  year={2022},
  publisher={Nature Publishing Group UK London}
}

@article{parthasarathy2024simclrplausible3,
  title={Layerwise complexity-matched learning yields an improved model of cortical area V2},
  author={Parthasarathy, Nikhil and H{\'e}naff, Olivier J and Simoncelli, Eero P},
  journal={ArXiv},
  pages={arXiv--2312},
  year={2024}
}

@article{loshchilov2017adamw,
  title={Decoupled weight decay regularization},
  author={Loshchilov, Ilya and Hutter, Frank},
  journal={arXiv preprint arXiv:1711.05101},
  year={2017}
}

@article{krizhevsky2012alexnet,
  title={Imagenet classification with deep convolutional neural networks},
  author={Krizhevsky, Alex and Sutskever, Ilya and Hinton, Geoffrey E},
  journal={Advances in neural information processing systems},
  volume={25},
  year={2012}
}

@inproceedings{stojanov2019incremental,
  title={Incremental object learning from contiguous views},
  author={Stojanov, Stefan and Mishra, Samarth and Thai, Ngoc Anh and Dhanda, Nikhil and Humayun, Ahmad and Yu, Chen and Smith, Linda B and Rehg, James M},
  booktitle={Proceedings of the IEEE/CVF conference on computer vision and pattern recognition},
  pages={8777--8786},
  year={2019}
}

@article{zhang2019selfrebuttal,
  title={A self validation network for object-level human attention estimation},
  author={Zhang, Zehua and Yu, Chen and Crandall, David},
  journal={Advances in Neural Information Processing Systems},
  volume={32},
  year={2019}
}

@inproceedings{stojanov2021using,
  title={Using shape to categorize: Low-shot learning with an explicit shape bias},
  author={Stojanov, Stefan and Thai, Anh and Rehg, James M},
  booktitle={Proceedings of the IEEE/CVF conference on computer vision and pattern recognition},
  pages={1798--1808},
  year={2021}
}

@inproceedings{tan2025prototypePAPN,
  title={Prototype-based Contrastive Learning with Stage-wise Progressive Augmentation for Self-Supervised Fine-Grained Learning},
  author={Tan, Baofeng and Wei, Xiu-Shen and Zhao, Lin},
  booktitle={Proceedings of the IEEE/CVF International Conference on Computer Vision},
  pages={4125--4134},
  year={2025}
}

@inproceedings{hou2023learnMADAug,
  title={When to learn what: Model-adaptive data augmentation curriculum},
  author={Hou, Chengkai and Zhang, Jieyu and Zhou, Tianyi},
  booktitle={Proceedings of the IEEE/CVF International Conference on Computer Vision},
  pages={1717--1728},
  year={2023}
}

@article{lu2025adoptingDVD,
  title={Adopting a human developmental visual diet yields robust, shape-based AI vision},
  author={Lu, Zejin and Thorat, Sushrut and Cichy, Radoslaw M and Kietzmann, Tim C},
  journal={arXiv preprint arXiv:2507.03168},
  year={2025}
}
}

\clearpage
\setcounter{page}{1}
\newpage
\onecolumn
{  
    \centering
    \Large
    \textbf{Supplementary Material for}\\
    \textbf{\thetitle}\\
}

\renewcommand{\thesection}{S\arabic{section}}
\renewcommand{\thefigure}{S\arabic{figure}}
\renewcommand{\thetable}{S\arabic{table}}
\setcounter{figure}{0}
\setcounter{section}{0}
\setcounter{table}{0}

\section{Implementation Details} \label{sec:temp}

\noindent \textbf{Training Details.} 

CATDiet integrates the staged curricula of CDiet and ADiet by interleaving their respective schedules, yielding an eight-stage training curriculum with stage lengths of [10, 6, 1, 5, 1, 2, 3, 2] epochs. Across these stages, we jointly manipulate the Gaussian blur standard deviation $\sigma$ and color-saturation ratio $s$. Specifically, we set $\sigma = [4, 3, 2, 2, 1, 1, 0, 0]$ (in pixels) and assign the saturation ranges for stages one through eight as (0.20, 0.36), (0.36, 0.52), (0.36, 0.52), (0.52, 0.68), (0.52, 0.68), (0.68, 0.84), (0.68, 0.84), and (0.84, 1.0), respectively.

We accelerate data loading using FFCV \cite{leclerc2023ffcv}. Training hyperparameters are selected via grid search for optimal performance. For SimCLR, we employ a staged temperature schedule; the temperature values $\tau$ for CDiet, ADiet, TDiet, CATDiet, and CombDiet are listed in \textbf{\cref{tab:simclrtemp}}.

\begin{table}[h]
\centering
\begin{tabular}{@{}lll@{}}
\toprule
         & $\tau$ value across stages                                            & stage durations         \\ \midrule
Cdiet    & {[}0.5, 0.4, 0.3, 0.2, 0.1{]}                           & {[}10, 7, 6, 5, 2{]}                           \\
Adiet    & {[}0.5, 0.4, 0.3, 0.2, 0.1{]}                           & {[}10, 6, 6, 3, 5{]}                           \\
Tdiet    & {[}0.1{]}                                           & {[}30{]}                                     \\
CATDiet  & {[}0.5, 0.45, 0.4, 0.35, 0.3, 0.2, 0.15, 0.1{]}     & {[}10, 6, 1, 5, 1, 2, 3, 2{]}                   \\
CombDiet & {[}0.5, 0.45, 0.4, 0.35, 0.3, 0.2, 0.15, 0.1, 0.1{]} & {[}10, 6, 1, 5, 1, 2, 3, 2, 70{]}                  \\ \bottomrule
\end{tabular}
\caption{\textbf{Temperature ($\tau$) schedules used for SimCLR under different visual diets.} Each row corresponds to one visual diet. The second column specifies the $\tau$ value used in each stage, and the third column provides the duration of each stage (in epochs).}
\vspace{-2mm}
\label{tab:simclrtemp}
\end{table}

For DINO, temperature is fixed throughout training; we use a student temperature of $\tau_s = 0.1$ and a teacher temperature of $\tau_t = 0.04$ for all visual diets. For momentum, we use the update schedule from \cite{caron2021emerging} for CATDiet and each individual diet (CDiet, ADiet and TDiet); in the two-phase CombDiet setting, the same update schedule is applied in Phase~1 and the momentum is reinitialized at the beginning of Phase~2 to adapt to SDiet.

\noindent \textbf{Calculation of mCE.} 
We compute the mean Corruption Error (mCE) following the protocol of \cite{hendrycks2019benchmarking}. ImageNet-C comprises 15 corruption types, each evaluated at 5 severity levels. For a classifier $f$, let $E^{f}_{s,c}$ denote the top-1 error under corruption type $c$ at severity level $s$. The aggregated error for corruption type $c$ is $E^{f}_{c} = \sum_{s=1}^{5} E^{f}_{s,c}$.

Because corruption types differ in difficulty, we normalize these errors by those of a baseline model, $E^{\text{baseline}}_{c}$. Following \cite{hendrycks2019benchmarking}, we use AlexNet \cite{krizhevsky2012alexnet} as the reference, yielding the normalized corruption error:
\[
CE^{f}_{c} = \frac{E^{f}_{c}}{E^{\text{AlexNet}}_{c}}
\]
The mCE is then obtained by averaging $CE^{f}_{c}$ over all 15 corruption types, providing an overall measure of corruption robustness for model $f$.

\noindent \textbf{Mapping Classes between IN and SAY.} 
The 15 IN \cite{ridnik2021in21k} classes used in this study and their corresponding SAY \cite{sullivan2021saycam} labels are listed below, where each IN label is followed by its corresponding SAY label in parentheses: n02124075 (cat), n09421951 (sand), n04399382 (plushanimal), n04590129 (window), n04239074 (door), n03125729 (crib), n02802426 (ball), n03201208 (table), n04344873 (couch), n03180011 (computer), n04099969 (chair), n03598930 (puzzle), n04285008 (car), n04204238 (basket), and n04462240 (toy).

\section{Ablation} \label{sec:suppablation}

\noindent \textbf{Schedule Ablation.}
For each diet, we vary stage durations to be either uniform or oppositely ordered from first to last stages, keeping the total number of epochs fixed. In \textbf{\cref{tab:schedule ablation}, left}, our default achieves the best Acc across diets. In \textbf{\cref{tab:schedule ablation}, right}, we also titrate the warm-up ratio in CombDiet (30\%, 50\%, 70\%) and find Acc is highest at 30\% (our default). 
\begin{table}[!h]
\centering
\small
\begin{tabular}{@{}lcccllc@{}}
\cmidrule(r){1-4} \cmidrule(l){6-7}
\textbf{Schedule} & \textbf{CDiet}         & \textbf{ADiet}         & \textbf{CATiet}        &  & \multicolumn{1}{l}{\textbf{Ratio}} & \textbf{CombDiet}      \\ \cmidrule(r){1-4} \cmidrule(l){6-7}
\textbf{Ours}     & \textbf{62.8} & \textbf{54.4} & \textbf{64.4} &  & \textbf{30\%}                              & \textbf{74.4} \\
Opposite & 59.8          & 52.1          & 63.1          &  & 50\%                              & 73.8          \\
Uniform  & 62.6          & 54.0          & 62.4          &  & 70\%                              & 73.1          \\ \cmidrule(r){1-4} \cmidrule(l){6-7}
\end{tabular}
\caption{Schedule ablation in Acc for SimCLR-ViT on CO3D-10.}
\label{tab:schedule ablation}
\end{table}

\noindent \textbf{Temporal Stride Ablation.}
We ablate the temporal sampling stride by halving or doubling the default setting used in CombDiet. As shown in \textbf{\cref{tab:temporal_stride_ablation}}, the performance remains stable across different strides, indicating that the method is robust to the choice of temporal sampling.

\begin{table}[!h]
\centering
\small

\begin{minipage}{0.48\textwidth}
\centering
\begin{tabular}{@{}lcc@{}}
\toprule
\textbf{Temporal Stride} & \textbf{Acc} & \textbf{mCE} \\ 
\midrule
\textbf{Ours} & 83.0 & 58.4 \\
0.5 x stride & 83.0 & 58.7 \\
2 x stride & 83.3 & 57.7 \\ 
\bottomrule
\end{tabular}
\caption{Temporal stride ablation (SimCLR-ResNet, CO3D-10).}
\label{tab:temporal_stride_ablation}
\end{minipage}
\hfill
\begin{minipage}{0.48\textwidth}
\centering
\begin{tabular}{@{}lcc@{}}
\toprule
\textbf{Pairing Strategy} & \textbf{Acc} & \textbf{mCE} \\ 
\midrule
\textbf{Ours} & 83.0 & 58.4 \\
Adjacent & 82.4 & 60.4 \\
Throughout & 70.6 & 73.0 \\ 
\bottomrule
\end{tabular}
\caption{Pairing strategy ablation (SimCLR-ResNet, CO3D-10).}
\label{tab:pairing_strategy_ablation}
\end{minipage}

\end{table}

\noindent \textbf{Pairing strategy ablation.}
We ablate whether artificial warping can replace temporally adjacent video frames for forming positive pairs.
Instead of pairing adjacent frames from the video, we consider two alternative strategies:
(1) \textbf{Adjacent}, where every other frame is replaced by a geometrically warped view and paired with its adjacent real frames to simulate local temporal continuity; and 
(2) \textbf{Throughout}, where a frame is paired only with warped views generated from different azimuth angles, without using any other real frames from the video. As shown in \textbf{\cref{tab:pairing_strategy_ablation}}, Adjacent performs close to Ours with a small but consistent drop, whereas Throughout performs substantially worse. This suggests that artificial warping can partially substitute adjacent frames and offers a practical trade-off between performance and storage efficiency. 
However, single-view warping fails to capture long-range temporal continuity, indicating that temporal continuity remains essential.

\section{Object recognition performance of CATDiet on other models}

\label{sec:otherablation}

\begin{figure}[hbt!]
    \centering
    \includegraphics[width=\textwidth]{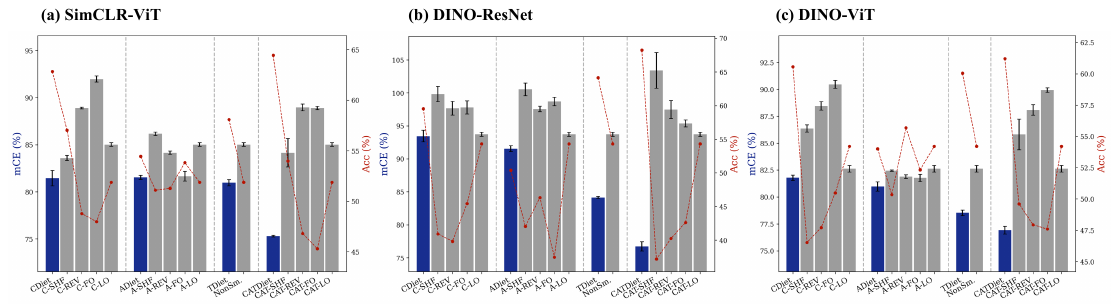}
    \caption{\textbf{Object recognition performance on CO3D (clean) and CO3D-C (corrupted) datasets for three SSL models pretrained on CATDiet and its individual diets.} Panels (a–c) show results for SimCLR-ViT, DINO-ResNet, and DINO-ViT, respectively. Within each panel, bars show mCE ($\downarrow$, left axis) for our diets (blue) and their baselines (gray); the dashed red line indicates Acc ($\uparrow$, right axis). Error bars reflect the standard error of mCE over three runs. Within each panel, the four groups correspond to different attributes of the proposed visual diets: Color, Acuity, Temporality, and their combination. 
    } 
    \label{fig:CATonothermodels}
\end{figure}

\noindent As shown in \textbf{\cref{fig:CATonothermodels}}, the conclusions in \textbf{\cref{sec:ablation}} consistently generalize across all SSL methods and backbones: Across all models (i.e., across all panels in \textbf{\cref{fig:CATonothermodels}}), each individual diet (CDiet, ADiet, TDiet) consistently achieves lower mCE than its corresponding baselines, demonstrating that developmentally-inspired visual diets promote robust representation learning across diverse SSL configurations. Moreover, the integrated CATDiet not only attains the lowest mCE among all diets but also exhibits larger performance gains in mCE and Acc over its baselines than any individual diet across all models, confirming that the synergistic benefit of integrating all developmentally-inspired diets generalize across other SSL methods and backbones.

\section{Object Recognition and Depth Order Classification Results of CombDiet pretrained on CO3D-10 and CO3D-50} \label{sec:combdietco3d}

\textbf{\cref{sec:combdietsay}} demonstrates that CombDiet improves model robustness and achieves superior performance on both object recognition and depth-order classification tasks. To assess whether these gains extend beyond SAY, we replicate the same experiments on CO3D. We begin with the 10-class CO3D subset used in the main text, and then evaluate scalability by repeating the experiments on an expanded 50-class version, which we denote CO3D-50.

\noindent \textbf{Datasets.}

\noindent \textbf{CO3D-50} \cite{reizenstein21co3d} and \textbf{CO3D-50-C}. We construct the \textbf{CO3D-50} benchmark using all 50 object categories in the CO3D dataset, sampling approximately 26{,}000 object instances for our experiments. Training and test splits are defined at the object-instance level within each category, and details of the frame-sampling procedure for both splits are provided in \textbf{\cref{sec:datasets}}. To evaluate robustness, we additionally construct a corrupted version of the CO3D-50 test set following the ImageNet-C protocol \cite{hendrycks2019benchmarking}, resulting in \textbf{CO3D-50-C}. For clarity, the 10-category subset used in the main text and its corrupted counterpart are referred to throughout as \textbf{CO3D-10} and \textbf{CO3D-10-C}, respectively.

\noindent\textbf{IN-10}, \textbf{IN-40}, \textbf{IN-10-C}, and \textbf{IN-40-C} \cite{ridnik2021in21k}.
For out-of-domain object recognition, we construct two ImageNet-21K subsets based on category overlap with CO3D-10 and CO3D-50.
For CO3D-10, the 10 overlapping ImageNet-21K categories form \textbf{IN-10}.
For CO3D-50, 40 categories align with ImageNet-21K, forming \textbf{IN-40}.
To evaluate robustness, we generate corrupted versions of the IN-10 and IN-40 test sets using the ImageNet-C protocol \cite{hendrycks2019benchmarking}, producing \textbf{IN-10-C} and \textbf{IN-40-C}.
Complete class mappings between CO3D-10 and IN-10, and between CO3D-50 and IN-40, are listed below, where each ImageNet label is followed by its corresponding CO3D class in parentheses:

The overlapping classes between IN-10 and CO3D-10 are:
n07930864 (cup), n04146614 (toybus), n03791053 (motorcycle), n04399382 (teddybear), n03809312 (toyplane), n04026813 (bicycle), n04099969 (chair), n03417042 (toytruck), n02958343 (car), and n02971579 (toytrain).

The overlapping classes between IN-40 and CO3D-50 are:
n07930864 (cup), n04146614 (toybus), n03791053 (motorcycle), n04399382 (teddybear), n04552348 (toyplane), n02835271 (bicycle), n04099969 (chair), n03417042 (toytruck), n04285008 (car), n04335435 (toytrain), n03483316 (hairdryer), n04522168 (vase), n04263257 (bowl), n04507155 (umbrella), n03891332 (parkingmeter), n04442312 (toaster), n04447861 (toilet), n02992529 (cellphone), n07248320 (book), n02802426 (ball), n04344873 (couch), n03983396 (bottle), n03991062 (plant), n03085013 (keyboard), n03793489 (mouse), n01608432 (kite), n07873807 (pizza), n02823750 (wineglass), n02769748 (backpack), n07747607 (orange), n03709823 (handbag), n07753592 (banana), n04404412 (tv), n04074963 (remote), n07742313 (apple), n07697537 (hotdog), n07714990 (broccoli), n03891251 (bench), n03642806 (laptop), and n03761084 (microwave).

\noindent \textbf{Detailed Settings.}
We evaluate CombDiet models pretrained on CO3D-10 and CO3D-50 across object recognition and depth-order classification. 
We summarize the specific training, linear probing, and test procedures below.

(1) \textit{Pretrained on CO3D-10}.
We pretrain four SSL models with CombDiet on the CO3D-10 training set.
For object recognition, we train 10-way linear probes on CO3D-10 and evaluate them on both clean CO3D-10 test images and the corrupted CO3D-10-C set.
To assess out-of-domain generalization, we train 10-way probes on IN-10 and evaluate on IN-10 and IN-10-C.
Depth perception is measured using a 2-way linear probe trained and tested on the 3D-PC Depth Order dataset.

(2) \textit{Pretrained on CO3D-50}.
We pretrain four SSL models with CombDiet on the CO3D-50 training set.
For object recognition, we train 50-way probes and evaluate them on CO3D-50 and CO3D-50-C.
For out-of-domain evaluation, we train 40-way probes on IN-40 and evaluate on IN-40 and IN-40-C.
Depth perception is again evaluated using a 2-way linear probe trained and tested on the 3D-PC dataset.

\noindent \textbf{Results.}
Following the presentation style in \textbf{\cref{sec:combdietsay}}, we report results in \textbf{\cref{tab:suppcombdietco3d}} for both CO3D-10 (Columns 3–7) and CO3D-50 (Columns 8–12).
On CO3D-10, CombDiet models achieve competitive or superior performance in Acc and dAcc, and outperform baselines by a substantial margin in mCE in most cases.
We observe similar trends on CO3D-50, which includes a much broader set of object categories.
These results demonstrate that the progressive ordering of developmental diets is essential: it substantially improves model robustness, whereas disrupting this order reduces performance to that of standard SSL training.

\begin{table}[h]
\resizebox{\textwidth}{!}{%
\begin{tabular}{cllcccccccccc}
\toprule
\multicolumn{1}{l}{}                    &                    &                                                                           & \multicolumn{5}{c}{\textbf{CO3D}}                                                                                                                                                                           & \multicolumn{5}{c}{\textbf{CO3D-50}}                                                                                                                                                                           \\ \cmidrule(lr){4-8} \cmidrule(lr){9-13}
\multicolumn{1}{l}{\textbf{}}           & \textbf{}          & \textbf{}                                                                 & \textbf{CO3D-10}                      & \textbf{CO3D-10-C}                    & \textbf{IN-10}                        & \textbf{IN-10-C}                      & \textbf{3D-PC}                        & \textbf{CO3D-50}                      & \textbf{CO3D-50-C}                    & \textbf{IN-40}                        & \textbf{IN-40-C}                      & \textbf{3D-PC}                        \\
\multicolumn{1}{l}{\textbf{}}           & 
\textbf{}          & \textbf{}                                                                 & \textbf{\cite{reizenstein21co3d}}                            & \textbf{\cite{hendrycks2019benchmarking}}                            & \textbf{\cite{ridnik2021in21k}}                            & \textbf{\cite{hendrycks2019benchmarking}}                            & \textbf{\cite{linsley20243d}}                            & \textbf{\cite{reizenstein21co3d}}                            & \textbf{\cite{hendrycks2019benchmarking}}                            & \textbf{\cite{ridnik2021in21k}}                            & \textbf{\cite{hendrycks2019benchmarking}}                            & \textbf{\cite{linsley20243d}}                            \\
\multicolumn{1}{l}{\textbf{}}           & \textbf{}          & \textbf{}                                                                 & \textbf{Acc}                          & \textbf{mCE}                          & \textbf{Acc}                          & \textbf{mCE}                          & \textbf{dAcc}                         & \textbf{Acc}                          & \textbf{mCE}                          & \textbf{Acc}                          & \textbf{mCE}                          & \textbf{dAcc}                         \\ \cmidrule{1-3} \cmidrule(lr){4-8} \cmidrule(lr){9-13}
\multicolumn{2}{c}{}                                         & \cellcolor[HTML]{F2F2F2}{\color[HTML]{333333} {\color{red}\textit{\textbf{Comb}}}{\textit{\textbf{Diet}}}}  
&\cellcolor[HTML]{F2F2F2}74.1 {\small $\pm$ 0.2}
& \cellcolor[HTML]{F2F2F2}68.0 {\small $\pm$ 0.5}
& \cellcolor[HTML]{F2F2F2}\textbf{80.1} {\small $\pm$ 0.7}
& \cellcolor[HTML]{F2F2F2}\textbf{57.4} {\small $\pm$ 0.4}
& \cellcolor[HTML]{F2F2F2}\textbf{76.8} {\small $\pm$ 1.2}
& \cellcolor[HTML]{F2F2F2}\textbf{75.7} {\small $\pm$ 0.2}
& \cellcolor[HTML]{F2F2F2}\textbf{76.8} {\small $\pm$ 0.2}
& \cellcolor[HTML]{F2F2F2}\textbf{65.0} {\small $\pm$ 0.3}
& \cellcolor[HTML]{F2F2F2}\textbf{76.1} {\small $\pm$ 0.3}
& \cellcolor[HTML]{F2F2F2}\textbf{74.1} {\small $\pm$ 1.5}\\
\multicolumn{2}{c}{}                                         & \textbf{SHF}                                                          
& 74.4 {\small $\pm$ 0.3}
& 70.0 {\small $\pm$ 0.3}
& 78.5 {\small $\pm$ 0.5}
& 61.6 {\small $\pm$ 0.4}
& 73.8 {\small $\pm$ 1.6}
& 70.9 {\small $\pm$ 0.1}
& 83.8 {\small $\pm$ 0.1}
& 61.3 {\small $\pm$ 0.2}
& 82.2 {\small $\pm$ 0.5}
& 68.5 {\small $\pm$ 1.4}   \\
\multicolumn{2}{c}{\multirow{-3}{*}{\makecell{\textbf{SimCLR-ViT}\\\textbf{\cite{chen2020simple} \cite{dosovitskiy2020image}}}}}    & \textbf{STD}                                                     
& \textbf{75.3} {\small $\pm$ 0.3}
& \textbf{67.8} {\small $\pm$ 0.2}
& 78.1 {\small $\pm$ 0.1}
& 58.3 {\small $\pm$ 0.1}
& 69.3 {\small $\pm$ 0.9}
& 73.0 {\small $\pm$ 0.7}
& 82.2 {\small $\pm$ 0.6}
& 62.1 {\small $\pm$ 0.3}
& 79.6 {\small $\pm$ 0.2}
& 65.3 {\small $\pm$ 1.3}               \\ \cmidrule{1-3} \cmidrule(lr){4-8} \cmidrule(lr){9-13}
\multicolumn{2}{c}{}                                         & \cellcolor[HTML]{F2F2F2}{\color[HTML]{333333} {\color{red}\textit{\textbf{Comb}}}{\textit{\textbf{Diet}}}} 
& \cellcolor[HTML]{F2F2F2}\textbf{83.0} {\small $\pm$ 0.2}
& \cellcolor[HTML]{F2F2F2}\textbf{58.5} {\small $\pm$ 0.3}
& \cellcolor[HTML]{F2F2F2}\textbf{88.1} {\small $\pm$ 0.4}
& \cellcolor[HTML]{F2F2F2}\textbf{45.8} {\small $\pm$ 0.5}
& \cellcolor[HTML]{F2F2F2}\textbf{71.5} {\small $\pm$ 1.3}
& \cellcolor[HTML]{F2F2F2}\textbf{82.4} {\small $\pm$ 0.0}
& \cellcolor[HTML]{F2F2F2}\textbf{70.3} {\small $\pm$ 0.1}
& \cellcolor[HTML]{F2F2F2}\textbf{75.6} {\small $\pm$ 0.2}
& \cellcolor[HTML]{F2F2F2}\textbf{69.3} {\small $\pm$ 0.2}
& \cellcolor[HTML]{F2F2F2}\textbf{75.9} {\small $\pm$ 1.7} \\
\multicolumn{2}{c}{}                                         & \textbf{SHF}                                                         
& 81.9 {\small $\pm$ 0.4}
& 62.9 {\small $\pm$ 1.0}
& 87.3 {\small $\pm$ 0.6}
& 50.4 {\small $\pm$ 0.4}
& 66.2 {\small $\pm$ 1.1}
& 79.8 {\small $\pm$ 0.1}
& 79.8 {\small $\pm$ 0.4}
& 73.8 {\small $\pm$ 0.6}
& 76.0 {\small $\pm$ 0.1}
& 68.8 {\small $\pm$ 1.7}                               \\
\multicolumn{2}{c}{\multirow{-3}{*}{\makecell{\textbf{SimCLR-ResNet}\\ \textbf{\cite{chen2020simple} \cite{he2016deep}}}}} & \textbf{STD}                                                  
& 81.0 {\small $\pm$ 0.1}
& 64.1 {\small $\pm$ 0.7}
& 87.4 {\small $\pm$ 0.5}
& 51.9 {\small $\pm$ 0.1}
& 65.0 {\small $\pm$ 1.1}
& 78.7 {\small $\pm$ 0.0}
& 84.0 {\small $\pm$ 0.4}
& 73.5 {\small $\pm$ 0.3}
& 76.9 {\small $\pm$ 0.4}
& 69.7 {\small $\pm$ 0.5}                 \\ \cmidrule{1-3} \cmidrule(lr){4-8} \cmidrule(lr){9-13}
\multicolumn{2}{c}{}                                         & \cellcolor[HTML]{F2F2F2}{\color[HTML]{333333} {\color{red}\textit{\textbf{Comb}}}{\textit{\textbf{Diet}}}} 
& \cellcolor[HTML]{F2F2F2}\textbf{77.7} {\small $\pm$ 0.7}
& \cellcolor[HTML]{F2F2F2}\textbf{62.4} {\small $\pm$ 0.2}
& \cellcolor[HTML]{F2F2F2}83.7 {\small $\pm$ 0.9}
& \cellcolor[HTML]{F2F2F2}\textbf{53.8} {\small $\pm$ 0.5}
& \cellcolor[HTML]{F2F2F2}\textbf{81.0} {\small $\pm$ 1.2}
& \cellcolor[HTML]{F2F2F2}80.2 {\small $\pm$ 0.1}
& \cellcolor[HTML]{F2F2F2}\textbf{67.8} {\small $\pm$ 0.1}
& \cellcolor[HTML]{F2F2F2}71.9 {\small $\pm$ 0.6}
& \cellcolor[HTML]{F2F2F2}\textbf{66.6} {\small $\pm$ 0.3}
& \cellcolor[HTML]{F2F2F2}\textbf{80.3} {\small $\pm$ 1.1} \\
\multicolumn{2}{c}{}                                         & \textbf{SHF}                                                         
& 73.7 {\small $\pm$ 0.1}
& 68.6 {\small $\pm$ 0.1}
& 80.1 {\small $\pm$ 1.3}
& 62.0 {\small $\pm$ 0.2}
& 78.0 {\small $\pm$ 0.9}
& 16.8 {\small $\pm$ 2.3}
& 115.9 {\small $\pm$ 2.4}
& 12.9 {\small $\pm$ 1.8}
& 116.8 {\small $\pm$ 2.5}
& 57.8 {\small $\pm$ 2.7}\\
\multicolumn{2}{c}{\multirow{-3}{*}{\makecell{\textbf{DINO-ViT}\\\textbf{\cite{caron2021emerging} \cite{dosovitskiy2020image}}}}}      & \textbf{STD}                  
& 76.8 {\small $\pm$ 0.6}
& 64.6 {\small $\pm$ 0.5}
& \textbf{84.3} {\small $\pm$ 0.8}
& 56.2 {\small $\pm$ 0.5}
& 80.0 {\small $\pm$ 1.1}
& \textbf{80.4} {\small $\pm$ 0.2}
& 72.2 {\small $\pm$ 0.7}
& \textbf{73.7} {\small $\pm$ 0.4}
& 68.6 {\small $\pm$ 0.4}
& 80.2 {\small $\pm$ 1.6}                    \\ \cmidrule{1-3} \cmidrule(lr){4-8} \cmidrule(lr){9-13}
\multicolumn{2}{c}{}                                         & \cellcolor[HTML]{F2F2F2}{\color[HTML]{333333} {\color{red}\textit{\textbf{Comb}}}{\textit{\textbf{Diet}}}} 
& \cellcolor[HTML]{F2F2F2}84.3 {\small $\pm$ 0.5}
& \cellcolor[HTML]{F2F2F2}\textbf{57.0} {\small $\pm$ 0.1}
& \cellcolor[HTML]{F2F2F2}89.1 {\small $\pm$ 0.3}
& \cellcolor[HTML]{F2F2F2}\textbf{49.3} {\small $\pm$ 0.5}
& \cellcolor[HTML]{F2F2F2}\textbf{73.1} {\small $\pm$ 1.2}
& \cellcolor[HTML]{F2F2F2}80.1 {\small $\pm$ 0.5}
& \cellcolor[HTML]{F2F2F2}\textbf{76.2} {\small $\pm$ 1.1}
& \cellcolor[HTML]{F2F2F2}76.2 {\small $\pm$ 0.6}
& \cellcolor[HTML]{F2F2F2}\textbf{74.5} {\small $\pm$ 0.9}
& \cellcolor[HTML]{F2F2F2}69.0 {\small $\pm$ 1.9}\\
\multicolumn{2}{c}{}                                         & \textbf{SHF}                                                         
& 76.8 {\small $\pm$ 0.6}
& 76.4 {\small $\pm$ 1.0}
& 83.1 {\small $\pm$ 1.4}
& 73.5 {\small $\pm$ 2.4}
& 64.9 {\small $\pm$ 0.8}
& 50.7 {\small $\pm$ 4.3}
& 106.2 {\small $\pm$ 2.0}
& 47.1 {\small $\pm$ 5.8}
& 104.7 {\small $\pm$ 2.3}
& 59.1 {\small $\pm$ 0.5}                           \\
\multicolumn{2}{c}{\multirow{-3}{*}{\makecell{\textbf{DINO-ResNet}\\ \textbf{\cite{caron2021emerging} \cite{he2016deep}}}}}   & \textbf{STD}                                                        
& \textbf{85.1} {\small $\pm$ 0.2}
& 63.6 {\small $\pm$ 0.5}
& \textbf{89.6} {\small $\pm$ 0.3}
& 59.8 {\small $\pm$ 0.5}
& 70.0 {\small $\pm$ 0.8}
& \textbf{85.0} {\small $\pm$ 0.1}
& 76.6 {\small $\pm$ 0.6}
& \textbf{80.3} {\small $\pm$ 0.4}
& 74.9 {\small $\pm$ 0.4}
& \textbf{76.3} {\small $\pm$ 1.2}                \\ \bottomrule
\end{tabular}
}
\caption{\textbf{Object Recognition and depth-order classification results of SSL models pretrained with CombDiet on CO3D-10 and CO3D-50 respectively.} Columns 3-7 denote results of models pretrained on CO3D-10, columns 8-12 denote results of models pretrained on CO3D-50. Each row corresponds to a specific [SSL]-[backbone] configuration (four in total). Within each row, three models are compared: CombDiet, SHF, and STD. Shaded rows highlight CombDiet performance. Values are reported in mean $\pm$ standard error of the mean (SEM) across three runs; best results are shown in \textbf{bold}. 
}
\label{tab:suppcombdietco3d}
\end{table}

\section{Additional Method Analysis} \label{sec:suppanalysis}

\begin{figure}[hbt!]
    \centering
    \includegraphics[width=\columnwidth]{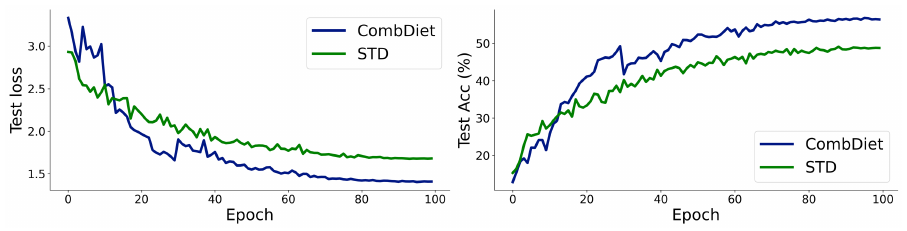}
    \vspace{-8mm}
    \caption{Online classification loss and Acc as a function of training epochs for models trained with STD (green) and CombDiet (blue) on SAY dataset.
    }
    \vspace{-4mm}
    \label{fig:convergence}
\end{figure}

\noindent \textbf{Training Convergence.}
To examine whether the models are sufficiently trained, we report online classification test loss and Acc over epochs. Both CombDiet and STD \textbf{(\cref{fig:convergence})} reach stable convergence; but CombDiet consistently achieves lower loss and higher Acc after saturation, indicating that our improved final performance of CombDiet over STD arises from better object representations rather than from early stopping.

\vspace{-4mm}

\noindent \textbf{Method Comparison.}
We compare our method with two prior works on curriculum learning,
including PAPN~\cite{tan2025prototypePAPN} and MADAug~\cite{hou2023learnMADAug}. 
PAPN targets fine-grained classification by aligning multiple augmented views of increasing strength across network depth. 
MADAug is developed for supervised classification, where augmentation policies are learned and adapt dynamically throughout training. 
Under the same training setup and evaluation protocols, CombDiet consistently outperforms both PAPN and MADAug, as shown in \textbf{\cref{tab: method comparison}}. These findings suggest that ecologically valid visual diets derived from infants are more effective than hand-crafted curricula for model training.

\begin{table}[!h]
\centering
\small

\begin{minipage}{0.48\textwidth}
\centering
\begin{tabular}{@{}lcc@{}}
\toprule
              & \multicolumn{1}{l}{\textbf{Acc}} & \multicolumn{1}{l}{\textbf{mCE}} \\ \midrule
\textbf{CombDiet} & \textbf{83.2}                              & \textbf{58.7}                                \\
PAPN \cite{tan2025prototypePAPN}          & 72.0                                         & 67.4                                         \\
MadAug \cite{hou2023learnMADAug}        & 51.7                                       & 101.6                                        \\ \bottomrule
\end{tabular}
\caption{Method Comparison for SimCLR-ResNet on CO3D-10. Best results are in bold.}
\label{tab: method comparison}
\end{minipage}
\hfill
\begin{minipage}{0.48\textwidth}
\centering
\begin{tabular}{@{}lcc@{}}
\toprule
                  & \textbf{Acc} & \textbf{mCE}  \\ \midrule
\textbf{CombDiet} & \textbf{66.0}  & \textbf{47.1} \\
\textbf{SHF}      & 65.7         & 61.5          \\
\textbf{STD}      & 63.5         & 65.2          \\ \bottomrule
\end{tabular}
\caption{Results for SimCLR-ResNet on TOYS. Best results are in bold.}
\label{tab: toys}
\end{minipage}

\end{table}

\noindent \textbf{Effectiveness on Synthetic Object-Centric Video Datasets.} Toys-200 and Toys4K~\cite{stojanov2019incremental,stojanov2021using} construct object-centric synthetic video datasets designed to approximate manipulable objects that infants encounter in early visual experience. We further evaluate our method on Toys4K, where we select 10 object classes and render multi-view images from diverse camera poses, denoted as \textbf{TOYS}. 
As shown in \textbf{\cref{tab: toys}}, CombDiet consistently achieves the best performance on TOYS, supporting the use of scalable synthetic object-centric environments for studying object learning under view continuity.

\noindent \textbf{Statistical Analysis.}
While a few settings in \textbf{\cref{tab:co3d-SAYCam}} and \textbf{\cref{tab:suppcombdietco3d}} favor STD mainly on Acc, CombDiet still achieves lower or comparable mCE, suggesting reduced shortcut learning. We perform t-tests based on the results reported in both tables. The p-values indicate statistically significant improvements of CombDiet over STD on mCE (p-value $<$ 0.05, \textbf{\cref{tab:pvalue}}).

\begin{table}[!h]
\centering
\small
\begin{tabular}{@{}lcccccc@{}}
\toprule
p-value         & SAY-C & IN-C & CO3D-10-C & IN-10-C & CO3D-50-C & IN-50-C \\ \midrule
CombDiet vs STD & $<0.001$     & 0.001    & 0.011      & 0.003   & 0.002     & 0.030  \\ \bottomrule
\end{tabular}
\caption{p-values from t-test on mCE (CombDiet vs. STD).
}
\label{tab:pvalue}
\end{table}

\section{Limitations} \label{sec:suppdiscussion}

There are several promising directions for future work. First, while we validate a sparse set of hyperparameter choices, exhaustive evaluation across all combinations remains infeasible due to the large search space. Future studies could efficiently expand the hyperparameter search to identify more optimal visual diets.
Second, although results on SAY suggest encouraging generalization, CombDiet is primarily designed for temporally coherent video clips featuring a single large, centrally located object. Its effectiveness in fully unconstrained, cluttered egocentric streams remains an open question.
Third, prior work \cite{zhang2019selfrebuttal} has explored active visual settings with human attention estimation for distant objects in cluttered scenes. Incorporating such attention mechanisms could further enhance model robustness and improve the ecological validity of infant visual diets.
Finally, while our current focus is on object recognition and depth perception, stress-testing SSL models trained with CombDiet on a broader range of downstream tasks—such as object detection, segmentation, and higher-level cognitive tasks \cite{zhang2017deep,zhang2018anticipating,sikarwar2023decoding,wang2025gazing,han2024flow,wang2023object,zhang2022look,jia2025seeing,zhang2017foveated}, remains an important avenue for future research.

\end{document}